\documentclass[10pt,twocolumn,letterpaper]{article}

\usepackage[pagenumbers]{cvpr} %

\usepackage{booktabs} %

\usepackage{amsmath}
\usepackage{algorithm}
\usepackage{algorithmic}
\usepackage[margin=1in]{geometry}
\usepackage{listings}
\usepackage{xcolor}
\usepackage{adjustbox}

\usepackage{pifont}
\newcommand{\xmark}{\ding{55}}%

\definecolor{cvprblue}{rgb}{0.21,0.49,0.74}
\usepackage[pagebackref,breaklinks,colorlinks,allcolors=cvprblue]{hyperref}

\title{\textbf{Explaining in Diffusion}: Explaining a Classifier Through Hierarchical Semantics with Text-to-Image Diffusion Models}
\author{
Tahira Kazimi\footnotemark[2] \quad
Ritika Allada\footnotemark[2] \quad
Pinar Yanardag \\
Virginia Tech \\
{\tt\small \{tahirakazimi, ritika88, pinary\}@vt.edu}
\\
{\href{https://explain-in-diffusion.github.io/}{\small \texttt{explain-in-diffusion.github.io}}}
}
\begin{document}

\twocolumn[{
\maketitle
\begin{center}
    \captionsetup{type=figure}
    \vspace{-2.4em}
\newcommand{\imwidth}{1\textwidth}

\begin{tabular}{@{}c@{}}
 
\parbox{\imwidth}{\includegraphics[width=\imwidth, ]{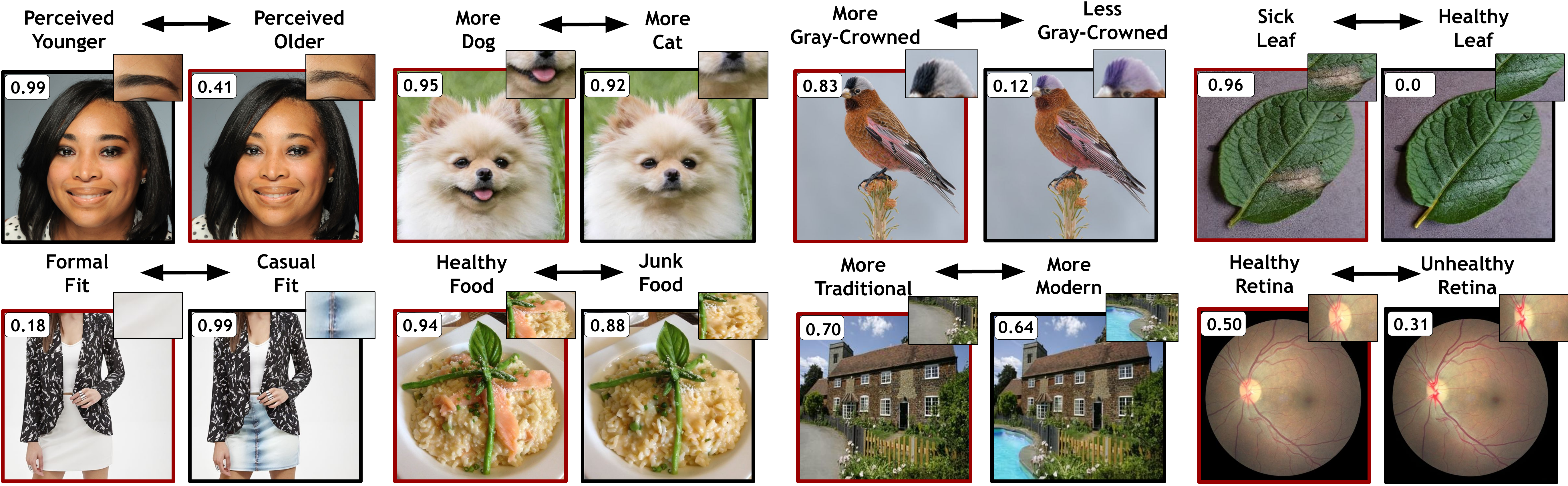}}
\\

\vspace{1em}
\end{tabular}
    \vspace{-3em}
    \captionof{figure}{\textit{DiffEx} explains the decisions of domain-specific classifiers by identifying the most influential semantics affecting their predictions. Classifier scores for each example are displayed in the top-left corner, demonstrating how classifier predictions change in response to the manipulation of different semantics (original images are shown with red borders). Our approach is capable of explaining classifiers that concentrate on individual concepts such as faces or animals (top row) as well as those that manage complex scenes involving multiple objects, such as a formal/casual fit in a fashion context (bottom row).} 
    \label{fig:teaser}
\end{center}
}]

\renewcommand{\thefootnote}{\fnsymbol{footnote}}
\footnotetext[2]{Joint first authors.}
\begin{abstract} 

Classifiers are important components in many computer vision tasks, serving as the foundational backbone of a wide variety of models employed across diverse applications. However, understanding the decision-making process of classifiers remains a significant challenge.  We propose DiffEx, a novel method that leverages the capabilities of text-to-image diffusion models to explain classifier decisions. Unlike traditional GAN-based explainability models, which are limited to simple, single-concept analyses and typically require training a new model for each classifier, our approach can explain classifiers that focus on single concepts (such as faces or animals) as well as those that handle complex scenes involving multiple concepts. DiffEx employs vision-language models to create a hierarchical list of semantics, allowing users to identify not only the overarching semantic influences on classifiers (e.g., the `beard' semantic in a facial classifier) but also their sub-types, such as `goatee' or `Balbo' beard.  Our experiments demonstrate that DiffEx is able to cover a significantly broader spectrum of semantics compared to its GAN counterparts, providing a hierarchical tool that delivers a more detailed and fine-grained understanding of classifier decisions.
\end{abstract}
\vspace{-1em} %
    
\section{Introduction}
\label{sec:intro}
Classifiers are fundamental to computer vision tasks, forming the backbone of many models used in a broad spectrum of applications \cite{krizhevsky2012imagenet, He_Zhang_Ren_Sun_2015, Simonyan_Zisserman_2014, lecun1998gradient, dosovitskiy2020image}. Their ability to generalize across tasks and adapt to new domains with minimal retraining makes them highly transferable, and thus they are employed extensively in fields such as healthcare \cite{9039644, Ozcift_Gulten_2011, Wiggins_Saad_Litt_Vachtsevanos_2008, pintelas2020explainable}, finance \cite{Sun_Li_2009, Peng_Wang_Kou_Shi_2011, Thakur_Kumar_2018}, security \cite{Ahsan_Gomes_Chowdhury_Nygard_2021, Leevy_Hancock_Zuech_Khoshgoftaar_2021, Katzir_Elovici_2018}, and autonomous systems \cite{Azouaoui_Chohra_2002, 7049114, Subramani2021}. Despite their versatility and widespread utility, understanding the decision-making process of classifiers remains a significant challenge \cite{Amgoud_2023, 10.5555/3304652.3304719, 10.1145/2939672.2939778, Zhang_2018_CVPR, zafar2021lack}. 
 This challenge stems largely from their ``black box" nature. As images traverse through the deep, interconnected layers of the network, the features used by the classifier to make a decision become increasingly abstract and challenging to interpret.
The lack of interpretable features in such models raises critical concerns, particularly in high-stakes environments such as medical diagnosis, where understanding the reasoning behind a model's prediction is crucial for ensuring trust, accountability, and informed decision-making \cite{rudin2022interpretable, zhang2021survey, chen2019looks, li2022interpretable, stiglic2020interpretability}. Explaining classifier decisions is crucial for enhancing the transparency and reliability of these models. Prior research \cite{lang2021explaining} has used generative adversarial networks (GANs)  \cite{goodfellow2020generative} to interpret classifier decisions by generating counterfactual examples that manipulate GAN latent semantics. These manipulations illustrate how changes in specific attributes, like the addition of the \textit{eyeglasses} semantic, impact classifier outputs. However, GANs are often limited to single domains, such as facial images, and typically require training a new model per classifier, which is resource and time-consuming. Moreover, in GAN-based methods, understanding which latent semantics affect classifier decisions often requires manual intervention to identify and interpret relevant features, such as recognizing that a discovered semantic controls the \textit{eyeglasses} attribute. This manual process is not only time-consuming but also less feasible in specialized fields like medicine, where identifying intricate attributes requires substantial domain expertise, making the approach impractical in critical scenarios.
 
This limitation highlights the need for more automated and versatile approaches to interpret classifier decisions. Text-to-Image (T2I) diffusion models \cite{stable-diffusion} emerge as a compelling alternative, widely recognized for their ability to generate high-quality images across various domains, which makes them a promising tool for explaining classifier decisions. These models offer the potential for a richer and more diverse set of semantic features compared to traditional methods. However, their ability to interpret and utilize latent space semantics remains limited in the context of diffusion models. Existing techniques for identifying meaningful semantics rely largely on supervised approaches \cite{brack2023sega,brack2024ledits++}, which require users to craft detailed text prompts to specify particular features for editing, such as \textit{mustache}. This process is labor-intensive and requires significant domain expertise, as users must carefully define prompts to edit the desired attributes. To make diffusion models effective for explaining classifier decisions, it is crucial to construct a comprehensive semantic corpus that covers large-scale semantics across various domains, thus minimizing reliance on manual input.   In this paper, we first employ Vision-Language Models (VLMs) \cite{OpenAI_Achiam_Adler_Agarwal_Ahmad_Akkaya_Aleman_Almeida_Altenschmidt_Altman_et} to extract a large-scale corpus covering domain-specific hierarchical semantics (see Fig. \ref{fig:bird-hierarchy}). Then, we introduce a training-free method, DiffEx, which leverages this hierarchical corpus and text-to-image diffusion models to explain the decision-making process of classifiers by identifying the most influential semantics. 
  Our method provides explanations for both coarse and fine-grained semantics. For example, it can recognize `beard' as a coarse semantic influencing age classification scores and also demonstrate how specific beard types (such as `Balbo' or `Anchor' beards) impact the classifier's scores. This hierarchical strategy provides users with an overview of the most significant semantics for a classifier, allowing them to dig deeper into particular fine-grained semantics that are essential for understanding classifier behavior.   Our qualitative and quantitative experiments reveal that DiffEx offers considerably richer and more comprehensive explanations for binary and multi-class classifiers across various domains, including facial features, retinal health, and plant pathology. Our contributions are as follows:

  \begin{figure}
    \centering
    \includegraphics[width=1.0\linewidth]{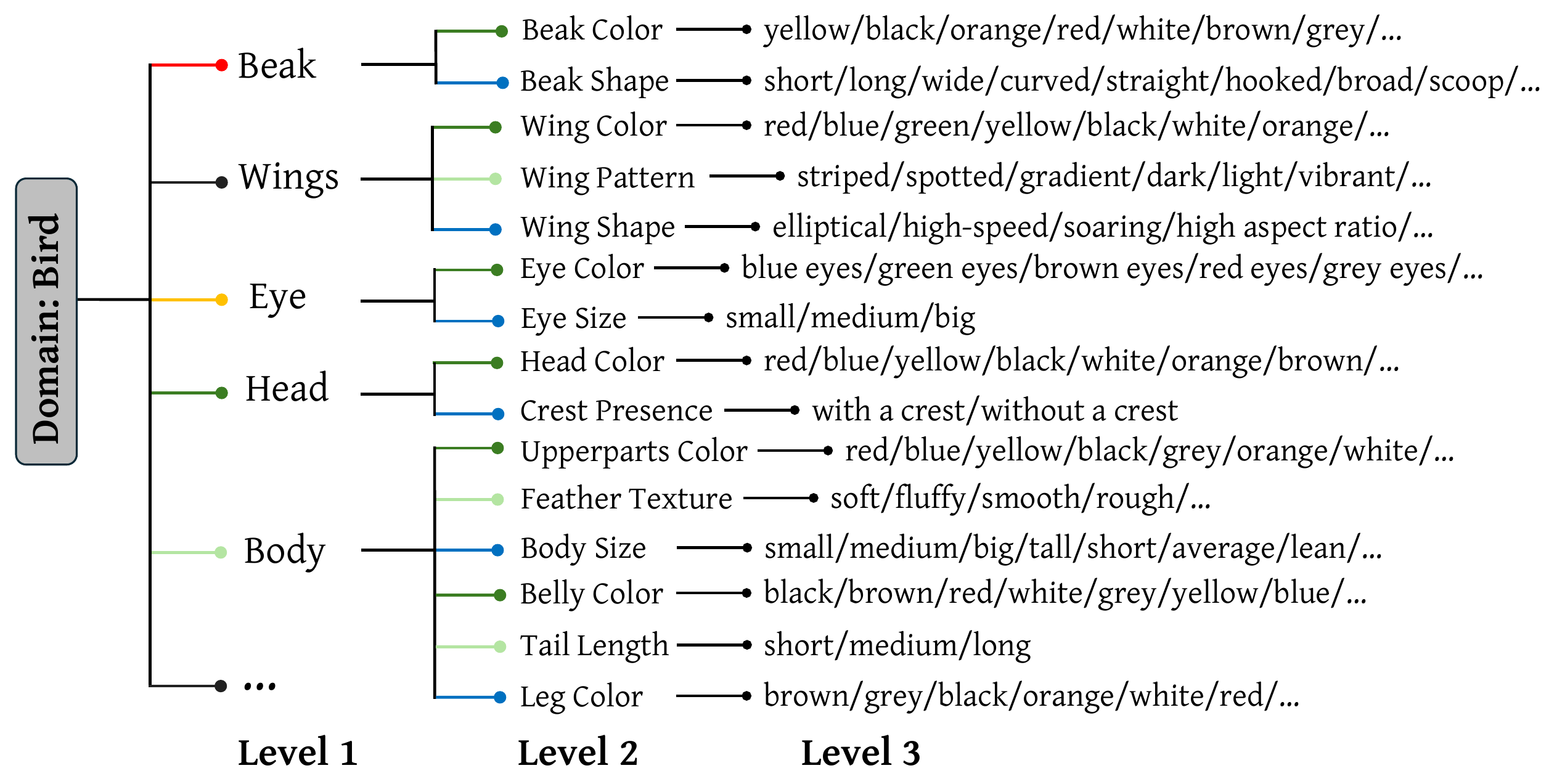}
    \caption{\textbf{Hierarchical List of Attributes for the Bird Domain.} We use VLMs to extract a hierarchical corpus of semantics within a given domain. This structured representation helps to illustrate how different attributes are grouped and their relationships within the broader domain, facilitating a better understanding of how each semantic contributes to the overall decision-making process of a classifier. 
 }
    \label{fig:bird-hierarchy}
\end{figure}

\begin{itemize}
    \item We propose \texttt{DiffEx}, a training-free approach using VLMs and T2I diffusion models to explain classifier decisions. To the best of our knowledge, this is the first hierarchical approach that explains classifier decisions. 
    \item Our method employs a VLM to develop a comprehensive semantic corpus that spans multiple domains in a hierarchical form. We make this corpus publicly available to  support future research.
    \item Unlike GAN-based methods, our approach can address classifiers that focus on single concepts (such as an `age' classifier analyzing a headshot of a person) and also extend to classifiers that assess complex scenes (such as a `Modern/Traditional Architecture' classifier evaluating entire scenes).
    \item We demonstrate that  \texttt{DiffEx}  offers more comprehensible and detailed explanations for classifiers in applications ranging from facial recognition to retinal health compared to prior approaches. Moreover, our method is adaptable and efficiently works with both binary and multi-classifiers.
\end{itemize}

\section{Related Work}

Traditional research focuses mainly on heat maps and patch-based extractions to explain classifier decisions. Specifically, class activation and saliency maps attempt to emulate human visual strategies by focusing on the image regions most relevant to a specific class \cite{simonyan2013deep, gradcam, saliency, XuVen, zeiler2014visualizing, zhou2016learning, srinivas2019full}. While these maps highlight the object or part of an image that most influences the classifier's decision, they fail to reveal which specific, fine-grained object attributes (e.g., color, pattern, or texture) impact the classification. Other approaches try to explain classifier outputs by analyzing extracted image patches \cite{Ghorbani_Wexler_Zou_Kim_2019, Yeh_Kim_Arik_Li_Pfister_Ravikumar_2022}; however, these methods can only reveal spatially localized attributes. 

\begin{figure*}[t!]
    \centering
    \includegraphics[width=1.0\linewidth]{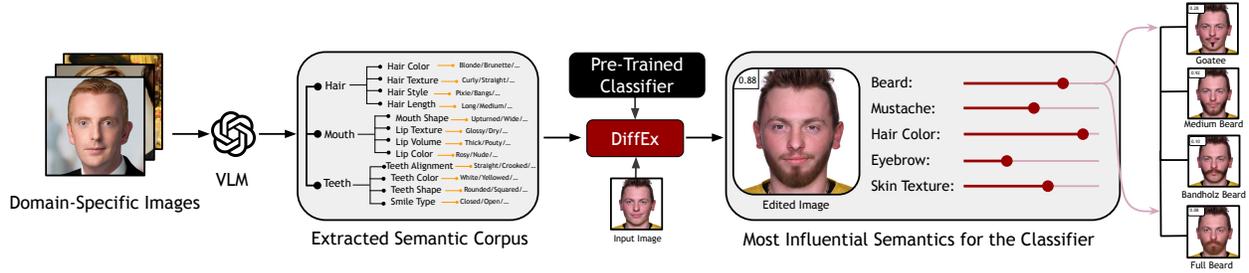}
    \caption{\textbf{An Overview of DiffEx.} Our pipeline processes a set of sample domain-specific images and a text prompt using a VLM to generate a hierarchical semantic corpus of attributes relevant to a specific domain. Based on this corpus, DiffEx identifies and ranks the most influential features affecting the classifier’s decisions, sorting them from most to least impactful (rightmost image). The hierarchical explanation of semantics (such as beard and its subtypes)  provides a fine-grained understanding of which features drive classifier outputs.}
    \label{fig:diffex}
\end{figure*}
Previous studies have leveraged generative models such as variational autoencoders (VAEs) \cite{goyal2019explaining}  or GANs  \cite{goetschalckx2019ganalyze, Sauer_Geiger_2021, Singla_Pollack_Chen_Batmanghelich_2020, Singla_Eslami_Pollack_Wallace_Batmanghelich_2022}. 
The most similar research to ours, StylEx \cite{lang2021explaining}, introduces a GAN-based approach to identify various attributes that influence a classifier's decisions. However, this method has several limitations. Firstly, it requires training a new GAN for each classifier, which can be resource-intensive and time-consuming. Secondly, each identified attribute requires manual labeling, which can require domain expertise. Lastly, StylEx only uncovers a limited range of semantic attributes, potentially overlooking others that might significantly affect a classifier’s scores. Additionally, a notable limitation of the StylEx method and similar GAN models is their focus on single concepts, such as \textit{human} or \textit{animal} faces, or individual objects like \textit{leaves}, rather than entire scenes. In contrast, diffusion models, known for their robust capability to generate complex and detailed entire scenes, offer a significant advantage. Our approach leverages the power of diffusion models to cover a wider spectrum of visual contexts, and enhances the versatility and applicability of our method across various classifiers that assess not only individual elements but also the interaction and composition of entire scenes.  

Recent approaches have begun using diffusion models to generate counterfactual examples.
One method utilizes shortcut learning to generate counterfactual images but fails to make semantically meaningful edits for certain attributes \cite{weng2025fast}.  
Another study explores modifying the diffusion process via adaptive parametrization and cone regularization to produce realistic counterfactual images; however, this approach depends on a robust model, which can be difficult to train \cite{Augustin_Boreiko_Croce_Hein_2022}.  \cite{jeanneret2022diffusion} explored counterfactual image generation, however their approach is computationally demanding and uses DDPM \cite{ho2020denoising} models trained on single domains.  As a result, it does not take advantage of large-scale latent diffusion models like Stable Diffusion, which can handle more complex scenes. Furthermore, even though some studies have leveraged diffusion models to generate realistic counterfactuals in high-stakes domains \cite{ilanchezian2023generating}, such as the medical field, they use a less efficient base model for the image generation process and focus predominantly on single-attribute modifications. Furthermore, to the best of our knowledge, there is no existing research that explores a hierarchical explanation of classifiers. Such an approach would systematically unpack the layers of influence that different semantic levels have on a classifier's scoring mechanism. This gap highlights a significant opportunity to enhance understanding by detailing how various semantic categories and their sub-types contribute to the decisions made by classifiers.
\section{Methodology}
\label{sec:method}

We first discuss the background on identifying attributes that influence classifier decisions through changes in logits. Following this, we introduce our method, which includes curating hierarchical attributes using vision-language models and a novel algorithm inspired by beam search to pinpoint attributes that affect classifier scores. Our pipeline is detailed in Fig. \ref{fig:diffex}. 

\subsection{Background}
\label{sec:stylex}
StylEx \cite{lang2021explaining} identifies semantics that meaningfully influence classifier decisions by ranking each attribute based on its impact on the classifier’s logit outputs. This ranking process aims to identify and select attributes that best explain the classifier’s behavior in a given context, such as understanding the factors influencing \textit{age} classification.

Given a semantic corpus \(\mathcal{S}\) and a set of \(N\) images to analyze the influence of various semantic attributes on classifier decisions, counterfactual images are first generated. For each original image \(x_i\), an edited version \(x'{_i} = g(x_{i}, s)\) is generated by applying a semantic attribute \(s \in \mathcal{S}\) through a transformation function \(g\). This transformation highlights the effect of each semantic attribute on a classifier's output. Specifically, the logit difference for each attribute is computed to measure how the classifier's score changes due to the presence of \(s\).

The influence of each attribute \(s\) is quantified as the average logit difference between the original and edited images across a set of sample images, defined as:
\begin{equation}\label{eq:stylexequation}
I(s) = \frac{1}{N} \sum_{i=1}^{N} |f(x'_i, y) - f(x_i, y)|
\end{equation}
where \(f(x, y)\) denotes the classifier’s logit score for target class \(y\) on image \(x\). Here, \(y\) represents the specific target class aimed to be explained, such as “age” or “gender.” This influence score \(I(s)\)  captures the average impact of each semantic attribute on classifier decisions by reflecting the logit score change due to attribute manipulation.

\subsection{Our Method}
 We first employ Vision-Language Models (VLMs) \cite{OpenAI_Achiam_Adler_Agarwal_Ahmad_Akkaya_Aleman_Almeida_Altenschmidt_Altman_et} to extract a large-scale corpus covering domain-specific hierarchical semantics. Then, we introduce a training-free method, DiffEx, which leverages this hierarchical corpus and text-to-image diffusion models to explain the decision-making process of classifiers by identifying the most influential semantics. 

\begin{algorithm}
\caption{DiffEx}
\begin{algorithmic}[1]
\REQUIRE Hierarchical structure $\mathcal{H}$ with semantic groups and features, beam width $B$, classifier or scoring function $f$, scoring threshold $\delta$
\ENSURE Optimal semantics maximizing $f$

\STATE Initialize $S \leftarrow \text{root-level groups in } \mathcal{H}$  \COMMENT{Initial candidate set at top-level groups}
\STATE Initialize beam $\mathcal{B} \leftarrow \emptyset$
\STATE Score each candidate $s \in S$ using the scoring function $f(s)$
\STATE Select top $B$ candidates with $f(b) \geq \delta$ and store in beam $\mathcal{B}$ \COMMENT{Apply thresholding to filter relevant candidates}
\WHILE{$S \neq \emptyset$}
    \STATE Initialize $S_{\text{next}} \leftarrow \emptyset$

    \FOR{each candidate $b \in \mathcal{B}$}
        \STATE Expand $b$ by adding sub-features from its next level in $\mathcal{H}$ to form new candidates
        \FOR{each new combination $b'$ generated from $b$}
            \IF{$f(b') > f(b)$}
                \STATE Add $b'$ to $S_{\text{next}}$
            \ENDIF
        \ENDFOR
    \ENDFOR

    \STATE Set $S \leftarrow S_{\text{next}}$  \COMMENT{Move to next level in hierarchy}
\ENDWHILE

\STATE Return highest-scoring combination from final $\mathcal{B}$ as the optimal joint semantic combination

\end{algorithmic}
\label{alg:joint}
\end{algorithm}
\subsubsection{VLM-Based Semantic Space}
\label{subsec:llm}
While the StylEx method outlined in Section \ref{sec:stylex} employs a logit-based approach to identify semantics, it requires manual labeling of each attribute extracted from the trained GAN model. Moreover, this method does not support the explanation of hierarchical attributes. Therefore, we first compile a large-scale set of semantics using VLMs. Given a domain \(d\), such as the `facial domain,' our objective is to identify a comprehensive set of domain-specific attributes from a collection of domain-specific images, denoted as \(N_d\). To accomplish this, we utilize a Vision-Language Model (VLM) \cite{OpenAI_Achiam_Adler_Agarwal_Ahmad_Akkaya_Aleman_Almeida_Altenschmidt_Altman_et} to extract a range of relevant features, represented by \(\mathcal{H}\). Employing in-context learning  \cite{brown2020language, su2022selective}, we prompt the VLM with a small set of images \(N_d\), along with a detailed task description and examples of desired outputs. This process allows us to generate a substantial semantic corpus of keywords, \(\mathcal{H}\), that captures the fine-grained attributes relevant to the domain.  The resulting corpus, \(\mathcal{H}\), comprises a comprehensive collection of keywords and phrases that covers the full spectrum of the domain's attributes. We leverage this rich dataset to provide a hierarchical explanation of classifier decisions, systematically explaining how different attributes influence outcomes.

\begin{figure*}[t!]
    \centering
    \includegraphics[width=0.97\linewidth]{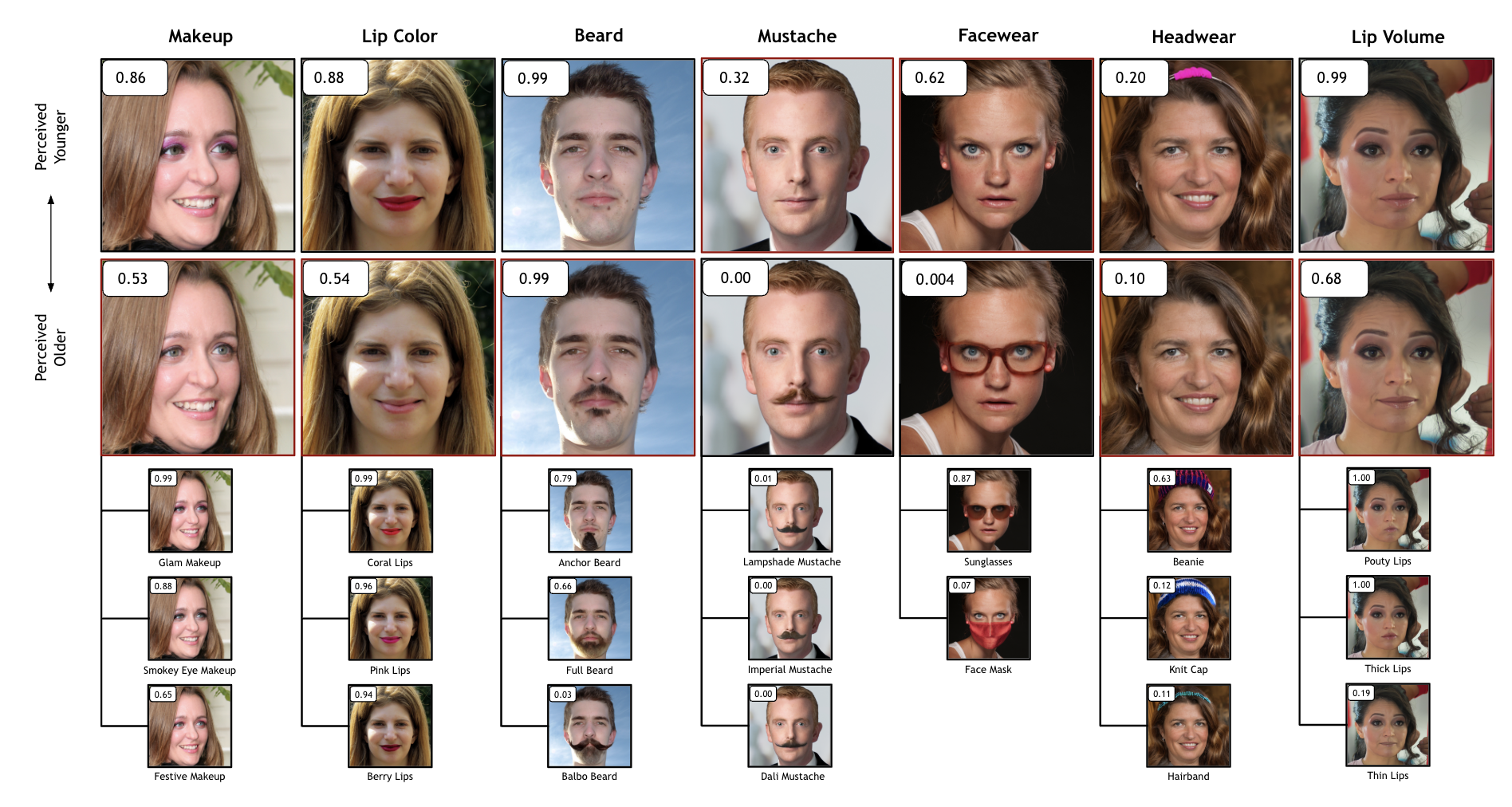}
    \caption{\textbf{Top-7 Discovered Facial Attributes for the Age Classifier.} DiffEx identifies key attributes and their top hierarchical subtypes for a perceived age classifier in the facial domain. For each attribute, the edited images and their respective subtypes are displayed in a hierarchical structure, outlined with a black border. The score for the ``young" label is shown in the top-left corner of each image. }
    \label{fig:face_main_fig}
\end{figure*}

\subsubsection{DiffEx}

\begin{figure*}[t!]
    \centering
    \includegraphics[width=1\linewidth]{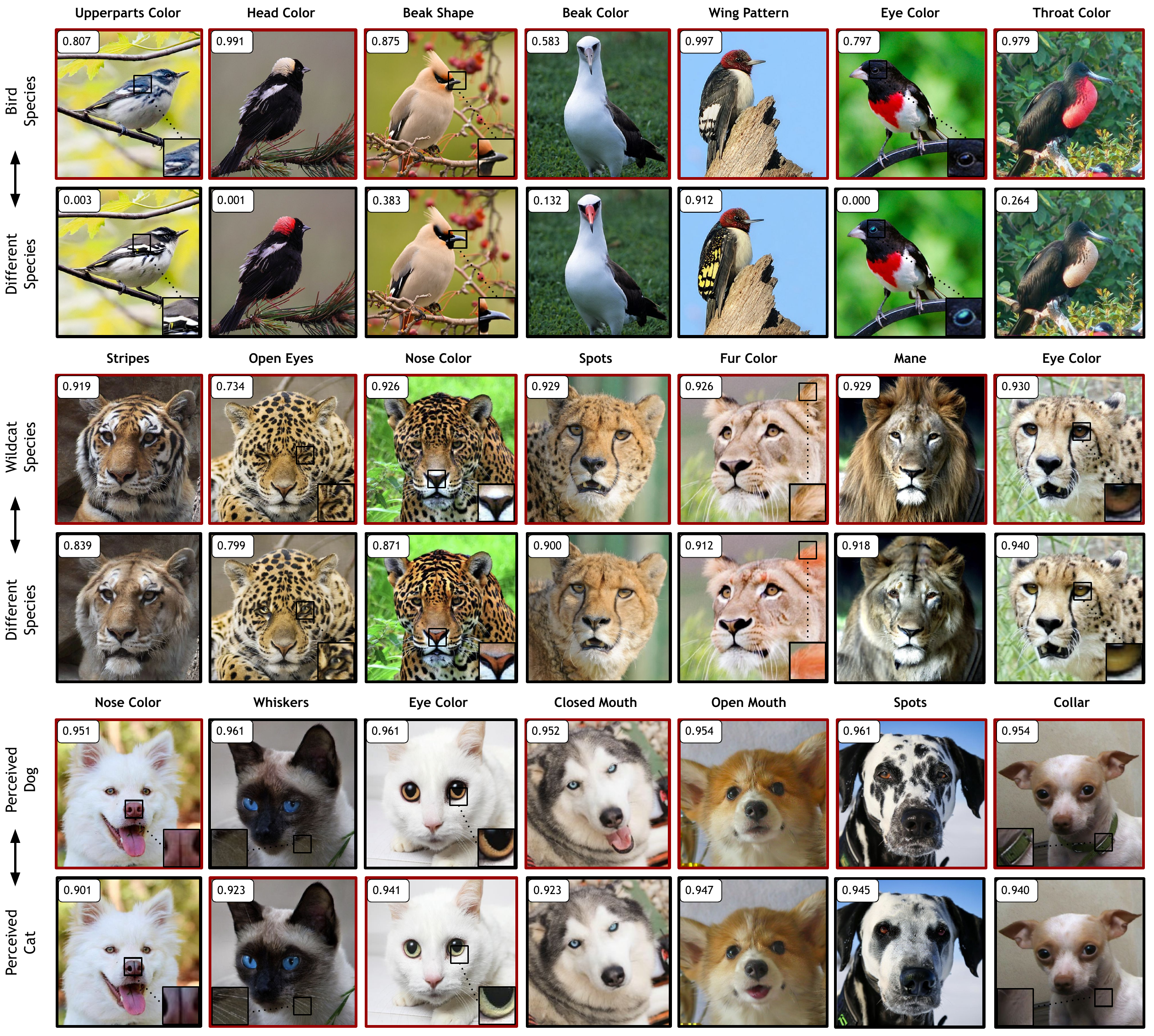}
    \caption{\textbf{Top-7 Discovered Attributes Across Different Animal Domains.} Our method successfully identifies key attributes for multiple domains, such as bird, wildcat, and pet species.  The original images are depicted with red borders while the edited images are depicted with black borders. For the pet species domain, we used a binary classifier and for the bird and wildcat species domains, we used a multi-classifier. The impact of each attribute on the classifier's score is shown in the top-left corner of each image. For attributes that caused subtle changes, we provided a zoomed-in view of the edit, displayed in the bottom left or right corner of the image.}
    \label{fig:birdwildcatpet}
\end{figure*}
Considering the extensive number of semantics identified using a VLM, it is computationally expensive to evaluate every possible semantic or combination of attributes. We introduce DiffEx, an efficient approach inspired by beam search to explain classifier decisions (see Algorithm \ref{alg:joint}). This method leverages our hierarchical semantic corpus to streamline the process. Our approach iteratively refines candidate attributes by expanding only the most impactful semantic paths at each hierarchical level, with each path's relevance guided by a scoring function that assesses the classifier’s response to each semantic feature. Formally, this scoring function calculates the average classifier $C$ score across $N$ sample images generated using each semantic attribute $s$, as defined in Eq. \eqref{eq:scoring_func}, where $g$ represents the generative diffusion model applied to introduce the attribute in each sample image:
\begin{equation}\label{eq:scoring_func}
    f(s) = \frac{1}{N} \sum_{i=1}^{N} C(g(x_{i},s))
\end{equation}
We begin with an initial candidate set $S$, which includes high-level groups from the semantic hierarchy (e.g., broader categories like `mouth features' or `eye features'). Each candidate in $S$ is evaluated by the scoring function $f(s)$, which quantifies the influence of the candidate on the classifier's output. Only the top $B$ candidates that surpass a predefined score threshold $\delta$ are retained, setting a beam width that limits the search to the most impactful candidates.
For each candidate in the beam, the algorithm proceeds by expanding to the next hierarchical level, incorporating more specific sub-features (e.g., for ``mouth features," sub-features such as ``beard" and ``mustache" are included). For each expanded candidate $b'$, the scoring function in Eq. \eqref{eq:scoring_func} is re-evaluated. Only candidates with a score exceeding that of their parent $f(b') >  f(b) $ are retained in the next candidate set $S_{next}$, ensuring that only those additions that yield significant incremental impact are added.
This process of expansion, scoring, and filtering continues iteratively, moving from general to more specific semantic attributes at each hierarchical level. By dynamically adjusting the candidate set based on the beam width $B$ and incremental scoring threshold $\delta$, our method remains computationally efficient, focusing on high-impact combinations rather than exhaustively evaluating all possible attribute pairings. For generating counterfactual images, we utilize an off-the-shelf editing tool for diffusion models, Ledits++ \cite{brack2024ledits++}. However, our approach is versatile enough to accommodate any editing method, allowing for flexibility in application and integration with different tools.

\begin{figure*}[t!]
    \centering
    \includegraphics[width=0.99\linewidth]{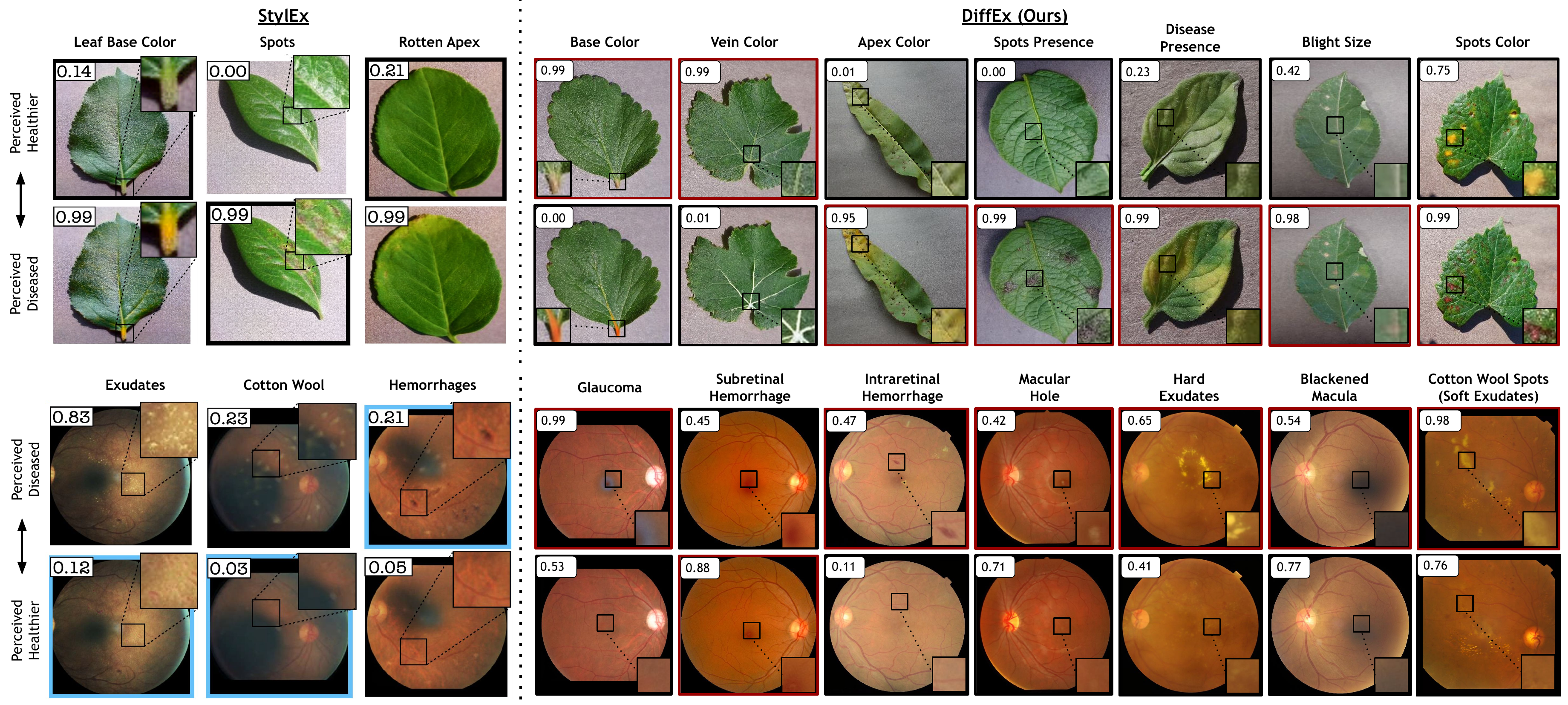}
    \caption{\textbf{Visual Comparison of Key Attributes Identified by StylEx and DiffEx in the Plant Health and Retinal Disease Domains.} This figure illustrates the enhanced capability of our method (DiffEx) in identifying a broader set of significant attributes compared to StylEx within the plant health and retinal disease domains. DiffEx successfully uncovers more detailed and diagnostically relevant features, such as  ``leaf vein color" and ``macular hole," which provide deeper insights into leaf and retina health. In contrast, StylEx primarily identifies general attributes like ``leaf base color" and ``exudates." For DiffEx, images with black borders represent the counterfactual images, while those with red borders represent the original images. For StylEx, images with blue or black borders are counterfactuals. For a comprehensive comparison of the top attributes discovered by StylEx and DiffEx across various domains, please refer to Table \ref{tab:comparisons}.}
    \label{fig:leafretina}
\end{figure*}

\section{Experiments}
\label{sec:experiments}
\subsection{Experimental Setup}

Our experiments utilize Ledits++ \cite{brack2024ledits++}  and Stable Diffusion XL (SDXL) \cite{podell2023sdxl}  for generating counterfactual images. Twenty-five time steps are omitted to boost computational efficiency while preserving the quality of edits. The edit threshold, a hyperparameter that dictates the global application scope of edits, is adjusted based on each domain. We test our method across diverse classifiers to evaluate its effectiveness in explaining model behavior through semantic influence. Specifically, we utilize classifiers trained on facial attributes data (i.e. age and gender classifiers) \cite{ffhq} as well plant health \cite{plantvillage}, retinal disease \cite{eyepacs}, bird species \cite{cub2011}, wildcat/pet species \cite{afhq}, fashion \cite{liu2016deepfashion}, places \cite{zhou2017places}, and food data \cite{bossard2014food}. 
The experiments for the face and plant health domains utilize a CLIP-based classifier \cite{radford2015unsupervised} to evaluate and interpret edits, while experiments within the bird, wildcat, pet, food, fashion, and places domains use CNN-based classifiers. These CNN classifiers were built on the EfficientNet \cite{tan2020efficientnetrethinkingmodelscaling} architecture and achieved an accuracy of over 95 percent on their test sets. For the retinal disease domain, we utilized the FLAIR model \cite{silva2025foundation}, which is based on a pre-trained vision-language model, to classify various retina scans. 
 
\begin{figure*}[t!]
    \centering
    \includegraphics[width=0.8\linewidth]{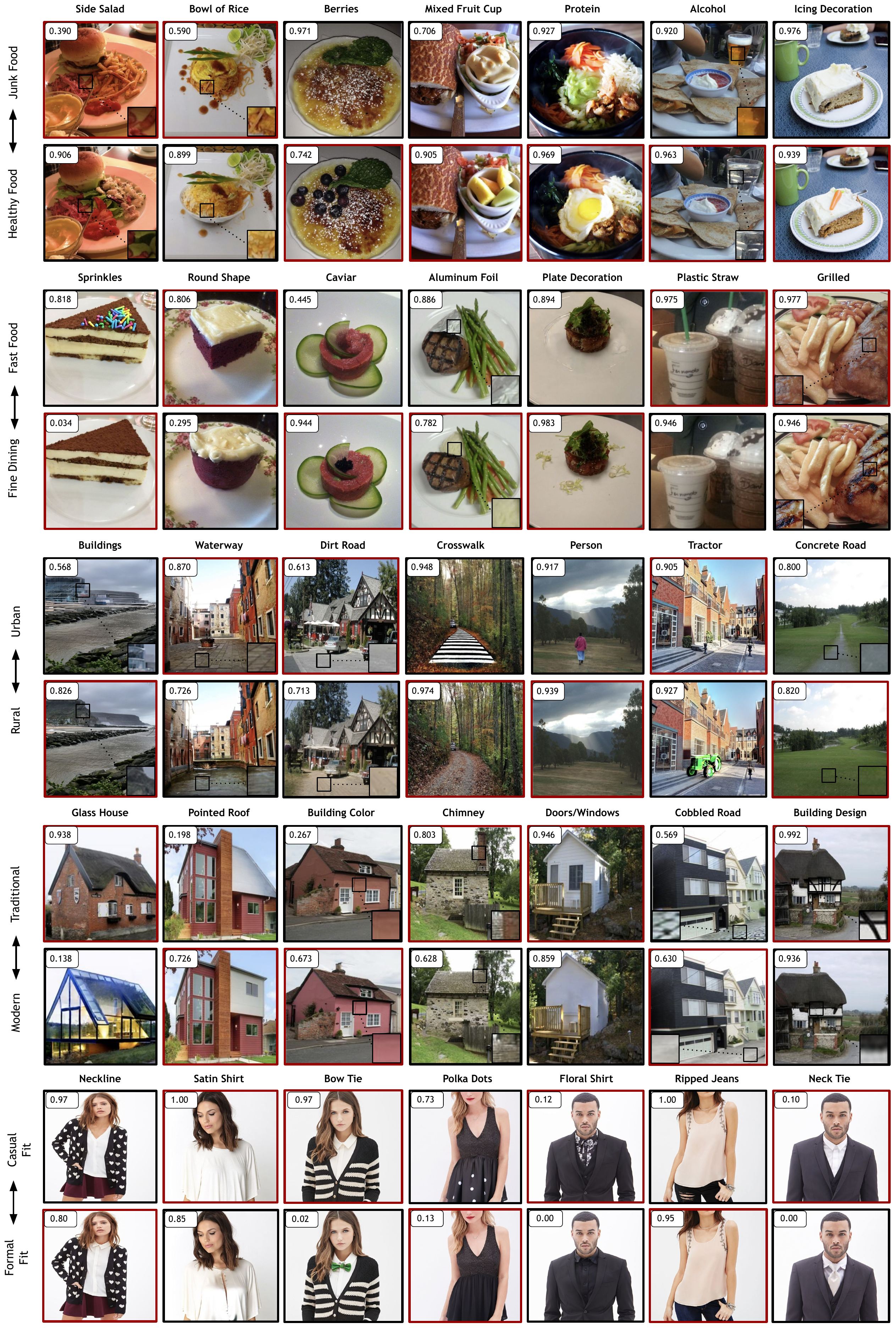}
    \caption{\textbf{Top-7 Attributes Discovered by DiffEx for Multi-Object Domains involving Complex Scenes.} Compared to existing methods, DiffEx was used to identify meaningful attributes for multi-object domains, such as ``food," ``places," and ``fashion." By identifying semantic features that make food appear ``healthier" or part of ``fine dining," as well as attributes that give a place a more ``rural" or ``modern" feel, DiffEx can serve as a powerful tool for understanding and modifying perceptions through targeted edits. As with the other examples in this paper, images with black borders indicate the original, unedited versions, while those with red borders represent the edited versions. For attributes with subtle changes, a zoomed-in view of the modification is shown in the image's bottom left or right corner.}
    \label{fig:foodbuildingimages}
\end{figure*}
\subsection{Qualitative Experiments}
\subsubsection{Explaining Classifiers with Single Attributes}
In the facial domain, we used a classifier to analyze fine-grained semantic features influencing the person's perceived age, while in other domains, we examined classifiers for species identification, leaf health assessment, retinal disease detection, food categorization, place perception, and clothing type.

\noindent \textbf{Face Domain:}  In the facial domain, as illustrated in Fig. \ref{fig:face_main_fig}, DiffEx identifies and ranks key attributes impacting age classification. For example, features such as ``makeup styles," ``lip volume," and ``accessories" (ex.``hairbands") are associated with perceived youthfulness, whereas attributes like ``facial hair" and ``eyeglasses" are linked with perceived older age.  Notably, the eyeglasses attribute consistently reduces the classifier's score for the ``young" label, reflecting its association with older demographics. Additionally, DiffEx uncovers hierarchical attribute structures, demonstrating how subcategories within a feature can have varying effects on classifier outcomes. For instance, as shown in Fig. \ref{fig:face_main_fig}, different beard styles impact the perceived age differently (e.g. a ``Balbo beard" significantly increases the age classification score more than a ``full beard"). Additional examples of hierarchical explanations of age classifiers are detailed in the appendix (\ref{fig:face_attr}).

\noindent \textbf{Animal Domain:}  Furthermore, DiffEx explains classifiers in a variety of animal types. Fig. \ref{fig:birdwildcatpet} highlights the top-7 most influential attributes in the bird, wildcat, and pet domains where our method was able to identify fine-grained semantics to explain classifier behavior.  For example, in the Wildcat classifier, attributes like \textit{stripes} and \textit{manes} are critical in distinguishing wildcats from other species. In contrast, in the Cat/Dog classifier, features such as \textit{whiskers, mouth position (open or closed)}, and \textit{spots} are among the key differentiators. Explaining the workings of these classifiers also reveals potential biases. For instance, the presence of a \textit{collar} significantly increases the likelihood of an animal being classified as a dog. This bias may stem from the training dataset where images of dogs more frequently featured collars compared to those of cats.

\noindent \textbf{Medical and Plant Health Domains:}  Fig. \ref{fig:leafretina} demonstrates the results for retinal disease and plant health domains. For the retinal disease domain, we use the FLAIR model to classify images of diseased and healthy retinal fundus scans. This model utilizes detailed domain expert knowledge descriptions, such as ``no hemorrhages, microaneurysms, or exudates” compared to general descriptions like ``no diabetic retinopathy" to aid in its classification. For the plant health domain, the CLIP classifier we use looks at features such as the presence of spots, fungus, or discoloration to make its decision. 

\noindent \textbf{Multi-Object Domains involving Complex Scenes:}   
Unlike traditional GAN-based methods that primarily focus on single-object scenarios like a cropped face, DiffEx extends its utility by providing a list of relevant features for domains encompassing multiple objects, such as places, food, and fashion. Classifiers within these domains often evaluate multiple elements simultaneously. For example, a fashion classifier might determine the `Formal or Casual Fit' of an outfit by considering various components such as hairstyle, top, and bottom. Similarly, an architectural classifier assessing whether a building appears `Urban or Rural' may base its evaluation on not just the structure itself but also its exteriors and surroundings. Explaining the decision-making process of such classifiers is crucial, as they analyze multiple objects within a single image, adding complexity to their interpretative frameworks. Fig. \ref{fig:foodbuildingimages} illustrates the top-7 attributes identified by DiffEx for the food, place, and fashion domains. Specifically, for the places domain, DiffEx identifies the top-7 attributes for the ``urban" vs. ``rural" and ``modern" vs. ``traditional" classifiers, while for the food domain, it does so for the ``healthy" vs. ``junk food" and ``fine dining" vs. ``fast food" classifiers. Fig. \ref{fig:foodbuildingimages} also demonstrates the top-7 attributes for ``casual" vs. ``formal" apparel classifier discovered by our method.

DiffEx is able to uncover interesting observations across various domains.  For example, for the food domain, DiffEx was able to discover that removing ``caviar” or a ``plate decoration” from the image made the food appear to be more perceived as fast food compared to fine dining. For the places domain, DiffEx was able to find that adding a ``tractor” to an image or a ``dirt road” made the place seem more rural than urban. In the fashion domain, the style of a neckline significantly influences whether an outfit is classified as ``formal" or ``casual." For instance, a V-neckline is often associated with a more casual look, whereas a classic boat neckline is generally perceived as more formal. Since DiffEx leverages text-to-image diffusion models, it can handle classifiers that interpret complex scenes involving multiple objects and provide detailed explanations for their decisions. This capability is essential for advancing understanding classifer decisions in diverse domains.

\paragraph{Explaining Classifiers with Joint Attributes}

\begin{table*}[htbp]
    \vspace*{\fill} 
    \centering
    \resizebox{\textwidth}{!}{
    \begin{tabular}{|l|l|l|l|l|l|l|l|l|l|l|l|}
        \hline
        \multicolumn{2}{|l|}{\textbf{Face (Age)}} & \multicolumn{2}{l|}{\textbf{Bird (Species)}} & \multicolumn{2}{l|}{\textbf{Leaves (Health)}} & \multicolumn{2}{l|}{\textbf{Retina Scans (Disease)}} & \multicolumn{2}{l|}{\textbf{Wildcat (Species)}} & \multicolumn{2}{l|}{\textbf{Pet (Cat/Dog)}} \\
        \hline
        \textbf{StylEx} & \textbf{Ours} & \textbf{StylEx} & \textbf{Ours} & \textbf{StylEx} & \textbf{Ours} & \textbf{StylEx} & \textbf{Ours} & \textbf{StylEx} & \textbf{Ours} & \textbf{StylEx} & \textbf{Ours} \\
        \hline
        Skin Pigmentation & Eyebrow & Belly Color & Upperparts Color & Base Leaf Color & Base Color & Exudates & Glaucoma & Spots & Stripes & Open Mouth & Nose Color \\
        Eyebrow Thickness & Makeup & Upperparts Color & Head Color & Apex Color & Vein Color & Cotton Wool Spots & Subretinal Hemorrhage & Black Tear Mark & Open Eyes & Closed Mouth & Whiskers \\
        Eyeglasses & Mustache Type & Wing Pattern & Beak Shape & Spots & Apex Color & Hemorrhages & Intraretinal Hemorrhage & Eye Shape + Size & Nose Color & Eye Shape & Eye Color \\
        Hair Color & Teeth & Beak Color & Beak Color & Blight & Spots Presence & Clustered Exudates & Macular Hole & \xmark & Spots & Dropped Ears & Closed Mouth \\
        Lip Thickness + Position & Lip Volume & Head Color & Wing Pattern & Halos & Disease Presence & \xmark & Hard Exudates & \xmark & Fur Color & Pointed Ears & Open Mouth \\
        Bangs & Lip Color & Breast Color & Eye Color & \xmark & Blight Size & \xmark & Blackened Macula & \xmark & Mane & Eye Circumference & Spots \\
        Eye Makeup & Eyelash & \xmark & Throat Color & \xmark & Leaf Texture & \xmark & Soft Exudates & \xmark & Eye Color & \xmark & Collar \\
        Facial Hair Color & Beard Type & \xmark & Wing Color & \xmark & Spots Color & \xmark & Retinal Drusen & \xmark & Tongue & \xmark & Pointed Ears \\
        \xmark & Facewear & \xmark & Crest Presence & \xmark & Discoloration & \xmark & Optic Disc Hemorrhage & \xmark & Pupil Size & \xmark & Mouth Color \\
        \xmark & Headwear & \xmark & Feather Texture & \xmark & Leaf Orientation & \xmark & Cataract & \xmark & Whiskers & \xmark & Fur Pattern \\
        \hline
    \end{tabular}
    }
    \caption{\textbf{Comparison of Top Attributes Across Different Domains and Classifiers.} The table above contains a list of the top attributes discovered by DiffEx (Ours) vs. StylEx. The \xmark~ in the table indicates attributes that were not mentioned in StylEx. It is also important to note that ``cotton wool spots" and ``soft exudates" refer to the same condition within the retinal disease domain.}
    \label{tab:comparisons}
    \vspace*{\fill} 
\end{table*}

\begin{figure}
    \centering
    \includegraphics[width=1.0\linewidth]{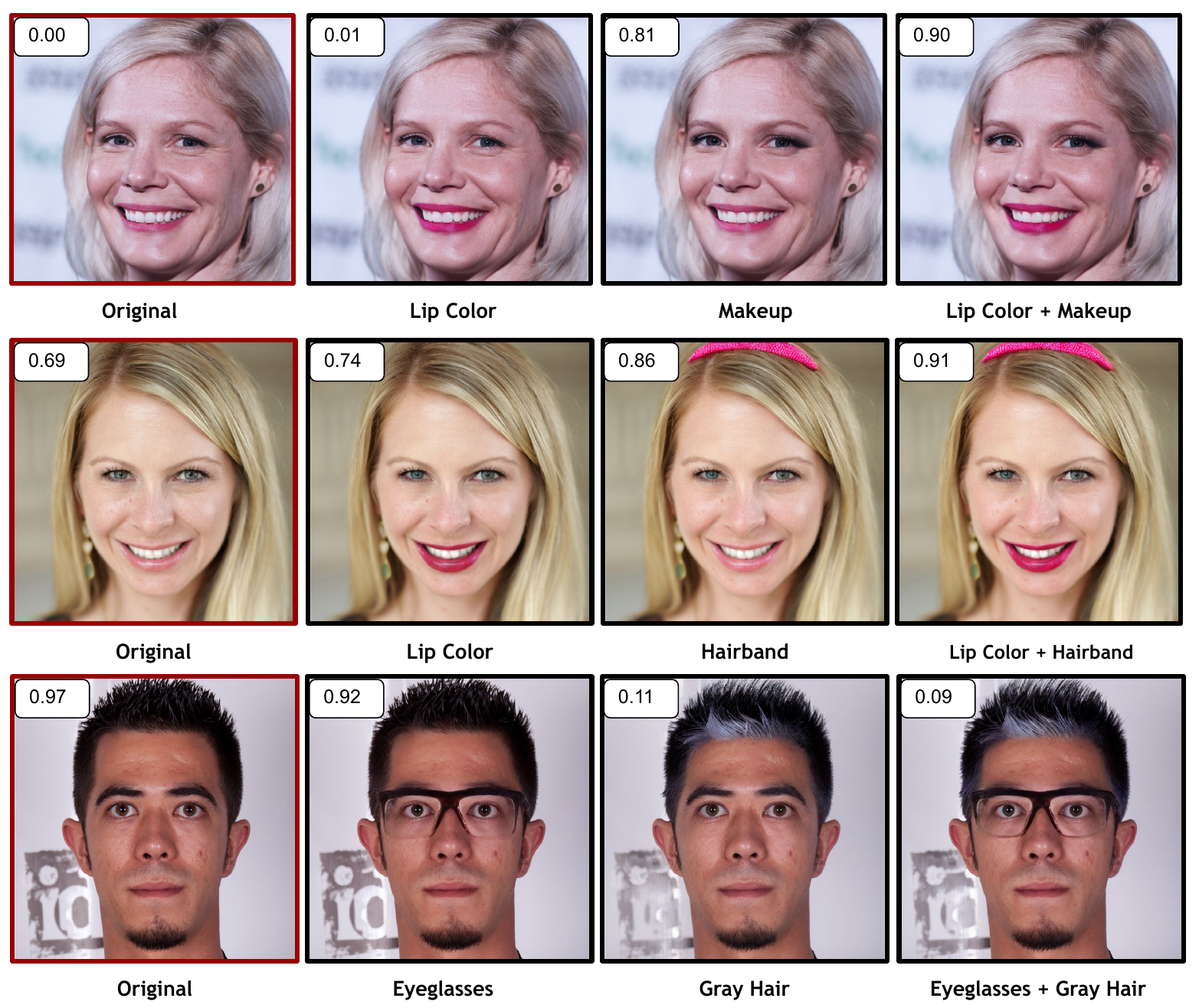}
    \caption{\textbf{Impact of Joint Attributes on Classifier Decisions.} DiffEx effectively explains classifier decisions by analyzing the combined influence of joint attributes. The score indicating the perceived age (``younger" label) is displayed in the top-left corner of each image, highlighting how specific attribute pairings can amplify or alter classifier outputs. }
    \label{fig:joint_edit}
\end{figure}

Our method goes beyond analyzing individual attributes by identifying attribute combinations that collectively improve classifier interpretation. While single attributes may have a minimal impact on logits, their combined effect can substantially influence the classifier’s output, uncovering subtle interactions that shape decision-making. For example, as shown in Fig. \ref{fig:joint_edit}, individual changes in ``lip color" or ``eye makeup" result in minimal score changes, but when modified together, they make the subject appear significantly younger. Similarly, adding a ``hairband" or changing the ``lip color" alone has little impact, but their combination shifts the classifier score markedly toward a younger age. On the other hand, pairing ``gray hair" with ``eyeglasses" strongly increases the perceived age, more so than each feature individually.
This analysis highlights how classifiers may respond more robustly to specific attribute combinations, providing deeper insights into feature interactions. Additional examples of joint attributes for the age, bird species, and plant health classifiers are provided in the appendix (\ref{fig:joint_face_sup}, \ref{fig:joint_leaf}, \ref{fig:joint_bird}).
\paragraph{Extending DiffEx for Multi-Classifier Analysis}
We adapt DiffEx for multi-classifier applications to uncover semantic attributes essential for tasks like retinal disease classification (Fig. \ref{fig:leafretina}) and bird and wildcat species identification (Fig. \ref{fig:birdwildcatpet}). This approach highlights key features such as ``upperparts color" and ``beak shape" for bird species, ``stripes" and ``nose color" for wildcat species, and types of hemorrhages and exudates for retinal conditions.
The results in Fig. \ref{fig:leafretina} for the retinal diseases domain and Fig. \ref{fig:birdwildcatpet} for the bird/wildcat species domain demonstrate that modifying these attributes significantly alters classifier scores, impacting assessments for species identification and disease likelihood.  
\subsubsection{Qualitative Comparison}

\begin{table*}[t!]
\centering
\resizebox{\textwidth}{!}{
\begin{tabular}{c c c c c c c c c c}
\hline
Method & Crest Presence & Beak Shape & Throat Color & Feather Texture & Eye Color & Beak Color & Head Color & Upperparts Color & Avg. Correct Response \\ \hline
Grad-CAM & 36\% & 50\% & 56\% & 35\% & 47\% & 65\% & 59\% & 76\% & 53\% \\ \hline
StylEx & 68\% & 85\% & 79\% & 82\% & 74\% & 68\% & 91\% & 65\% & 76.5\% \\ \hline
\textbf{DiffEx (Ours)} & \textbf{88\%} & \textbf{91\%} & \textbf{88\%} & \textbf{91\%} & \textbf{82\%} & \textbf{82\%} & \textbf{97\%} & \textbf{88\%} & \textbf{88.4\%} \\ \hline
\end{tabular}
}
\caption{\textbf{Comparison with Other Explainability Methods.} The table above displays the percentage of correct attribute selections for the bird class, as chosen by users when viewing outputs from different explainability methods. It also includes the average percentage of correct responses across all attributes for each method. As shown, for each attribute presented, the majority of users identified the correct attribute when viewing the output generated by DiffEx.}
\label{tab:comps}
\end{table*}
\begin{table}[t!]
    \centering
    \renewcommand{\arraystretch}{1}
    \begin{tabular}{p{2.7cm} c c}
    \hline
    \textbf{Rating} & \textbf{Bird Domain} & \textbf{Face Domain} \\ \hline
    \textbf{Edit Quality} & 3.386 $\pm$ 0.223 & 3.659 $\pm$ 0.248 \\ \hline
    \textbf{Disentanglement} & 3.163 $\pm$ 0.197 & 3.204 $\pm$ 0.213\\ \hline
\end{tabular}
    \caption{\textbf{Edit Quality and Disentanglement Ratings.} The table above provides the average edit quality and faithfulness ratings across different domains from User Study 1. The scoring is done on a scale from 1 to 5.}
    \label{tab:quality_faithfulness}
\end{table}

In Fig. \ref{fig:leafretina}, we present a visual comparison of the top attributes and classification scores identified by DiffEx against those identified by StylEx for the plant health and retinal disease domains. As illustrated, DiffEx successfully uncovers a broader set of semantically meaningful attributes compared to StylEx. For instance, DiffEx identifies detailed attributes such as ``leaf vein color" and ``spots color," in addition to the more general attributes found by StylEx, like ``leaf base color" and ``apex color." This expanded set of attributes enables a more detailed understanding of key diagnostic features, especially relevant in applications such as plant health assessment. Table \ref{tab:comparisons}  provides a comprehensive comparison of the top features identified by StylEx and our method, highlighting DiffEx's superior ability to uncover semantics that are crucial for the classifier's decision-making process. In particular, the table presents an extended list of features across domains such as faces, bird species, plant health, retina scans, wildcat species, and pet types, all of which influence a classifier’s score—further emphasizing DiffEx's ability to uncover fine-grained and contextually rich features. This highlights DiffEx's advantage in providing a more comprehensive understanding of classifier behavior by capturing both general and specific feature variations within a given category.

In addition to identifying common features found by StylEx, such as ``eyebrow thickness," ``makeup," and ``facial hair," DiffEx demonstrates a broader and more generalized ability to detect relevant features in the facial domain. Unlike StylEx, which tends to focus on specific attribute variations, such as ``facial hair color," DiffEx captures a wider range of feature types and their detailed effects. For example, DiffEx is capable of distinguishing between different beard shapes and understanding how each individual style influences the classifier’s decision, as shown in Fig. \ref{fig:face_main_fig}. While StylEx may highlight ``facial hair color" as a key feature, DiffEx’s approach covers the diversity within the category, such as differentiating between a ``full beard," ``Balbo beard," and ``anchor beard," and analyzing their respective impacts on perceived age.

\subsection{Quantitative Experiments}
\paragraph{Baselines} 
To quantitatively evaluate the effectiveness of DiffEx compared to other explainability methods, such as StylEx \cite{lang2021explaining} and Grad-CAM \cite{gradcam}, we conducted a series of comprehensive user studies to assess how easily participants could identify the relevant features extracted by our method.

\noindent \textbf{User Study 1: Visual Quality and Disentanglement} To evaluate the visual quality and disentanglement of the edited images for the face and bird domains, we conducted a user study with 50 participants on Prolific\footnote{Prolific, \url{https://www.prolific.com}}. Specifically, for each domain, we showed pairs of unedited and edited images for ten attributes and asked the users to assess whether the edited image contained the desired attribute and if the edit appeared to be disentangled. For each pair of images, we asked the users to rate the edit and disentanglement from one to five, with five representing the highest score. Our results indicate that our edits successfully reflected the intended attributes while minimizing any unrelated changes (see Table \ref{tab:quality_faithfulness}). Please refer to Appendix for more details about our user study. 

\noindent \textbf{User Study 2: Comparison with Grad-CAM and StylEx} To evaluate how effectively our method, DiffEx, explains various semantics within a specific domain compared to other explainability methods (Grad-CAM and StylEx), we conducted a user study with 35 participants on Prolific. We focused on images from the bird domain, generating three sets of three images per attribute: one original image, one edited image, and one image illustrating the explainability method. For Grad-CAM, participants were shown a heatmap overlay on the edited image. For StylEx, we displayed the edited image alone, as this method requires users to manually label the edits. For DiffEx, participants viewed the edited image along with the attribute automatically assigned by the VLM. We then asked users to select the attribute (e.g., ``beak color," ``crest presence") that best explained the edited image from four answer choices. Results from this study indicate that DiffEx significantly outperforms Grad-CAM and StylEx in explaining edited attributes in images (see Table \ref{tab:comps}). Please refer to Appendix for more details about our user study.

\section{Limitation}
\label{sec:limitation}
While our approach offers significant insights into classifier behavior through semantic edits, there are a few limitations to consider. First, since our method relies on semantics curated through VLMs, it is constrained by the quality and scope of the initial semantic corpus. This corpus may not fully capture all relevant or nuanced features, especially in specialized domains where unique or context-specific attributes are necessary for accurate interpretation. Another limitation is that since we are utilizing an off-the-shelf image editing model, such as Ledits++, to target specific attributes, some edits may be entangled (a general issue in image editing algorithms \cite{wu2023uncovering}), and might introduce confounding factors that could affect classifier scores. This is particularly important in high-stakes domains, such as medical imaging, where even minor unintended changes may impact interpretability.  Nevertheless,  our framework is flexible and can be improved with domain-specific semantic adjustments such as task-specific RAGs \cite{wu2024medical} or other editing methods \cite{dalva2023noiseclr, brack2023sega}.

\section{Conclusion}
\label{sec:conclusion}
In this work, we introduce DiffEx, a novel approach for explaining classifier decisions by utilizing semantic edits within diffusion models. By harnessing the power of vision language models, we curate a comprehensive, hierarchical semantic corpus across various domains and propose a novel algorithm inspired by beam search to filter and rank the most impactful features. DiffEx ranks these semantic features based on their influence on classifier logits, capturing both individual and joint attribute effects, which are crucial for understanding complex classifier behaviors. Through experiments conducted on a wide range of domains—including face, bird, and medical classifiers—we showcase the robustness and adaptability of our approach. This work provides a powerful tool for interpreting model decisions across diverse applications, promoting transparency, and building trust in AI-driven classification systems.

{
    \small
    \bibliographystyle{ieeenat_fullname}
    \bibliography{main}
}
\clearpage %
\setcounter{section}{0} %
\renewcommand{\thesection}{S\arabic{section}} %
\setcounter{figure}{0} %
\renewcommand{\thefigure}{S\arabic{figure}} %
\setcounter{table}{0} 
\renewcommand{\thetable}{\Alph{table}}

\clearpage
\setcounter{page}{1}
\maketitlesupplementary
\section{Overview}
\label{sec:rationale}
In this appendix, we present further details on our methodology, user studies, and experiments. Specifically, we include the prompt template used with our chosen vision-language model, GPT-4, as well as a more detailed hierarchy of attributes across various domains. We also share further insights from our user studies, including the questions asked and examples of images presented during the evaluations. Finally, we include additional examples of single- and joint-attribute editing and demonstrate how integrating the counterfactual images we generated into the training data of a classifier can improve its accuracy and robustness.
\section{GPT-4 Prompt Template}
To extract a list of potential attributes for each domain, we provided the prompt template in Table \ref{tab:gpttemplate} to GPT-4. A more detailed example of the text prompt for the face domain can be found in Table \ref{tab:gptface}.
\lstset{
    basicstyle=\ttfamily\small,
    breaklines=true,
    frame=single,
    backgroundcolor=\color{gray!10},
    keywordstyle=\color{blue},
    stringstyle=\color{green!60!black},
    commentstyle=\color{red!50!black},
    columns=fullflexible
}
\section{Hierarchy of Attributes}
\begin{figure}
    \centering
    \includegraphics[width=1.0\linewidth]{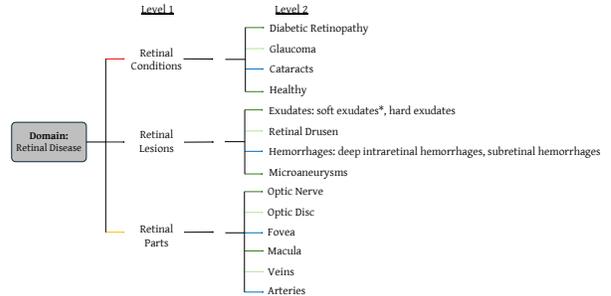}
    \caption{\textbf{Hierarchical List of Attributes for the Retinal Disease Domain.} The diagram above illustrates the hierarchical organization of various attributes within the retina scan domain, showing the levels to which they belong. Note: The asterisk next to ``soft exudates" denotes that they are also referred to as ``cotton wool spots," and the sub-categories under ``exudates" are part of the level 3 hierarchy.
 }
    \label{fig:retina-hierarchy}
\end{figure}

Figures \ref{fig:bird-hierarchy} and \ref{fig:retina-hierarchy} depict the hierarchical structures of various attributes for the bird and retinal disease domains respectively. Tables \ref{tab:attributes}, \ref{tab:attributes2} and \ref{tab:attributes3} show an extensive list of potential attributes for the face, plant health, and bird domains respectively. Level 1 attributes refer to ``broader" categories while Level 2 and Level 3 attributes refer to ``finer-grained" categories. It is important to note that the attributes listed in these figures represent only a subset of all the attributes provided by GPT-4.
Additionally, Table \ref{tab:retinaranks} features the top-10 attributes for the face, bird, plant health, and retinal disease domains, and their corresponding ranking scores.
\begin{table}[t!]
    \centering
    \renewcommand{\arraystretch}{1.2} %
    \setlength{\tabcolsep}{3pt} %
    \small %
    \begin{tabular}{p{2.5cm}@{}p{4.2cm}@{}c@{}}
        \hline
        \textbf{Domain} & \textbf{Top-10 Attributes} & \textbf{Score} \\ \hline
        \textbf{Face} 
        & \raggedright Eyebrow & 0.74 \\ 
        & \raggedright Makeup & 0.50 \\ 
        & \raggedright Mustache & 0.47 \\ 
        & \raggedright Teeth (Smile) & 0.44 \\ 
        & \raggedright Lip Volume & 0.37 \\ 
        & \raggedright Headwear & 0.33 \\ 
        & \raggedright Lip Color & 0.25 \\ 
        & \raggedright Beard & 0.15 \\ 
        & \raggedright  Facewear & 0.11 \\ 
        & \raggedright Hair & 0.10 \\
        \hline
        \textbf{Bird} 
        & \raggedright Upperparts Color & 0.55 \\ 
        & \raggedright Head Color & 0.55 \\ 
        & \raggedright Beak Shape & 0.38 \\ 
        & \raggedright Beak Color & 0.37 \\ 
        & \raggedright Wing Pattern & 0.29 \\ 
        & \raggedright Eye Color & 0.28 \\ 
        & \raggedright Throat Color & 0.27 \\ 
        & \raggedright Wing Color & 0.26 \\ 
        & \raggedright Crest Presence & 0.13 \\ 
        & \raggedright Feather Texture & 0.04 \\ \hline
        \textbf{Plant Health} 
        & \raggedright Leaf Base Color & 0.97 \\ 
        & \raggedright Leaf Vein Color & 0.91 \\ 
        & \raggedright Leaf Apex Color & 0.89 \\ 
        & \raggedright Leaf Spots & 0.84 \\ 
        & \raggedright Leaf Disease & 0.77 \\ 
        & \raggedright Leaf Blight Size & 0.16 \\ 
        & \raggedright Leaf Spots Color & 0.10 \\ 
        & \raggedright Leaf Texture & 0.07 \\ 
        & \raggedright Leaf Discoloration & 0.04 \\ 
        & \raggedright Leaf Orientation & 0.03 \\ \hline
        \textbf{Retinal Disease} 
        & \raggedright Glaucoma & 0.43 \\ 
        & \raggedright Subretinal Hemorrhage & 0.42 \\ 
        & \raggedright Intraretinal Hemorrhage & 0.35 \\ 
        & \raggedright Macular Hole & 0.33 \\ 
        & \raggedright Hard Exudates & 0.33 \\ 
        & \raggedright Blackened Macula & 0.23 \\ 
        & \raggedright Soft Exudates & 0.21 \\ 
        & \raggedright Retinal Drusen & 0.13 \\ 
        & \raggedright Optic Disc Hemorrhage & 0.05 \\ 
        & \raggedright Cataract & 0.04 \\ \hline
    \end{tabular}
    \caption{\textbf{Top-10 Attributes and their Respective Scores Across Various Domains.} The table above displays the top 10 attributes for the face, bird, plant health, and retinal disease domains, ranked from highest to lowest based on their scores. These scores were derived by calculating the average difference between the classification scores of the edited and unedited images.}
    \label{tab:retinaranks}
\end{table}
\section{Algorithmic Comparison}
The big-O algorithmic comparisons between StyleEx \cite{lang2021explaining} and DiffEx are approximated in Table \ref{tab:alg_tab}. Both algorithms perform at $O(n^3)$ in the worst case, where $n$ denotes the number of sample images and $s$ denotes the number of style vectors for the GAN-based approach and the set of semantics in the diffusion-based approach. StyleEx uses a greedy search approach to find the most relevant attribute. However, the average complexity of this algorithm is heavily dependent on the number of sample images and the length of the style vectors. If the number of images $n$ is much smaller than the length of style vector $s$, ($n << s$), then the complexity becomes $O(s^2)$. However, if $n >> s$, then the time complexity becomes $O(n^2)$. On the other hand, DiffEx is linear with respect to the number of semantics $s$, the number of sample images $n$, and the beam width $b$. For $n >> s$, DiffEx performs at $O(b \cdot n)$ whereas for $n << s$, the complexity becomes $O(b \cdot s)$. Furthermore, the beam search approach strikes a compromise between the efficiency of greedy search and optimality of exhaustive search \cite{zhang2021dive}. Therefore, beam search can produce more optimal outcomes with the help of its beam width logic compared to the greedy method. 
\begin{table}[t!]
    \centering
    \renewcommand{\arraystretch}{1}
    \begin{tabular}{p{2.3cm} c c c}
    \hline
    \textbf{Algorithm} & \textbf{Worst Case} & \textbf{Average Case} \\ \hline
    \textbf{StylEx} &  $O(n^3)$ & $O(n^2)$  \\ \hline
    \textbf{DiffEx (Ours)} & $O(s^2 \cdot n)$ & $O(b \cdot n \cdot s)$ \\ \hline
\end{tabular}
    \caption{\textbf{Big-O Comparisons Between Greedy-Based StylEx and Beam-Based DiffEx.} The table above provides estimations of the algorithmic efficiency of both methods. In StyleEx, $n$ represents the number of sample images and $s$ is the length of style coordinates. On the other hand, $s$ in DiffEx represents the hierarchical structure of keywords extracted from the VLM.}
    \label{tab:alg_tab}
\end{table}

\section{Quantitative Evaluation}

For our primary quantitative evaluation, we conducted two user studies to assess different aspects of our approach. \textbf{User Study 1} focused on evaluating edit quality and disentanglement, while \textbf{User Study 2} compared our method, DiffEx, against two explainability techniques: Grad-CAM and StylEx. 
We chose user studies as the main quantitative assessment because they directly evaluate the human-centric goals of our explainability method. Explainability is ultimately about making AI systems more interpretable and useful for humans, which user studies are well-suited to measure. There is also no universally accepted benchmark for explainability, and by focusing on user studies, we ensure that our evaluation captures real-world factors across diverse domains. 
Subsections \ref{subsec:us1} and \ref{subsec:us2} include some example questions from these user studies.
\subsection{User Study 1: Evaluating Edit Quality and Disentanglement}
\label{subsec:us1}
\begin{figure}
    \centering
    \includegraphics[width=1.0\linewidth]{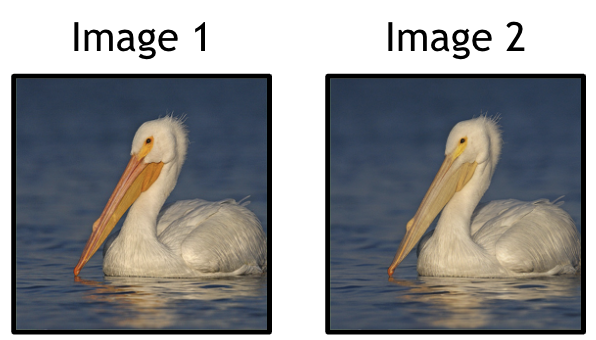}
    \caption{\textbf{Example of Original and Edited Image Comparison from User Study 1.} The image pair above serves as an example from the quantitative study on edit quality and disentanglement, specifically for the ``beak color" attribute. }
    \label{fig:bird-beak}
\end{figure}

In this study, participants were shown eight pairs of images per domain 
(specifically the ``face" and ``bird" domains). Each pair consisted of an edited image and its corresponding unedited version to highlight the change in a specific attribute. Participants were then asked to evaluate the edits by answering the following questions, which quantitatively assessed the quality and disentanglement of the modifications. For example, for the \textit{beak color} attribute in the bird domain, participants were shown a sample pair of images (as seen in Figure \ref{fig:bird-beak}) and asked the following questions:
\begin{itemize}
    \item \textbf{Edit Quality:}  
    \textit{Given the original image (image 1) and edited image (image 2), how likely do you think the modified image reflects the intended change (e.g., beak color)?}  \\
    \textbf{Scale:} 1 = Not Likely, 5 = Very Likely
    \item \textbf{Disentanglement:}  
    \textit{Given the original image (image 1) and edited image (image 2), how likely do you think the edited image is disentangled compared to the original image? Disentanglement means the modification performed only the desired edit (e.g., modifying the ``beak color" without altering unrelated areas).} \\
    \textbf{Scale:} 1 = Not Disentangled, 5 = Fully Disentangled
\end{itemize}
These questions enabled us to systematically evaluate the effectiveness of the edits in achieving the desired modifications while maintaining disentanglement. The results of the study are included in Table \ref{tab:quality_faithfulness} in the Quantitative Experiments section of our paper. %
\subsection{User Study 2: Comparisons with Grad-CAM and StylEx}
\label{subsec:us2}

\begin{figure}
    \centering
    \includegraphics[width=1.0\linewidth]{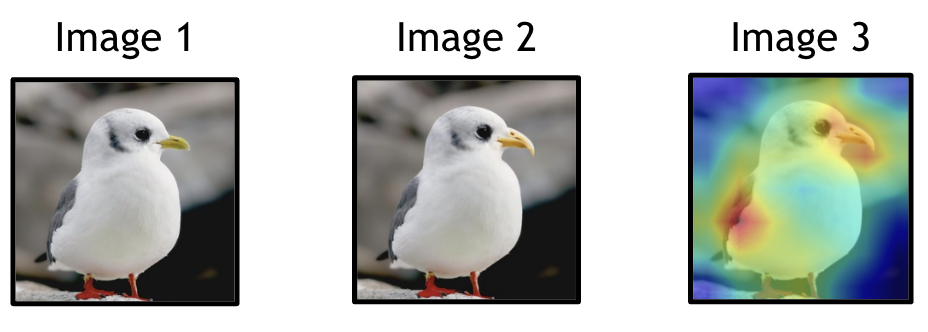}
    \caption{\textbf{Comparison of Original, Edited, and Grad-CAM Images from User Study 2.} Image 1 depicts the original image of a bird, Image 2 shows the edited version of the bird, and Image 3 illustrates the Grad-CAM explainability metric, highlighting the most important attribute(s) in the edited image. In this example, ``beak shape" was the attribute that was edited (as seen in image 2); however, Grad-CAM highlights both the bird's wing and beak, making it unclear which attribute is the primary focus of the image.}
    \label{fig:grad-cam-bird}
\end{figure}
To quantitatively evaluate our method, DiffEx, against other explainability metrics, we conducted another user study. Participants were presented with 3 sets of 3 images per attribute: the original image, an edited (counterfactual) image, and a third image explained using a comparable metric, such as Grad-CAM. For each set, participants were asked the following question, with modified answer choices and corresponding images: ``Given three images (image 1, image 2, and image 3), select the attribute that best describes the feature highlighted in image 3." A sample set of answer choices provided for the question accompanying Figure \ref{fig:grad-cam-bird} were:
\begin{enumerate}[label=\alph*.)]
    \item Feather Texture
    \item Beak Shape
    \item Beak Color
    \item Eye Color
\end{enumerate}
\section{Additional Experiments}
In this section, we present some further experiments that demonstrate the impact of single and joint attributes on the classifier's output. We also explore how training the classifier on counterfactual examples can enhance its robustness.
\subsection{Experiments with Single Attributes}
\begin{figure*}
    \centering
    \includegraphics[width=0.95\linewidth]{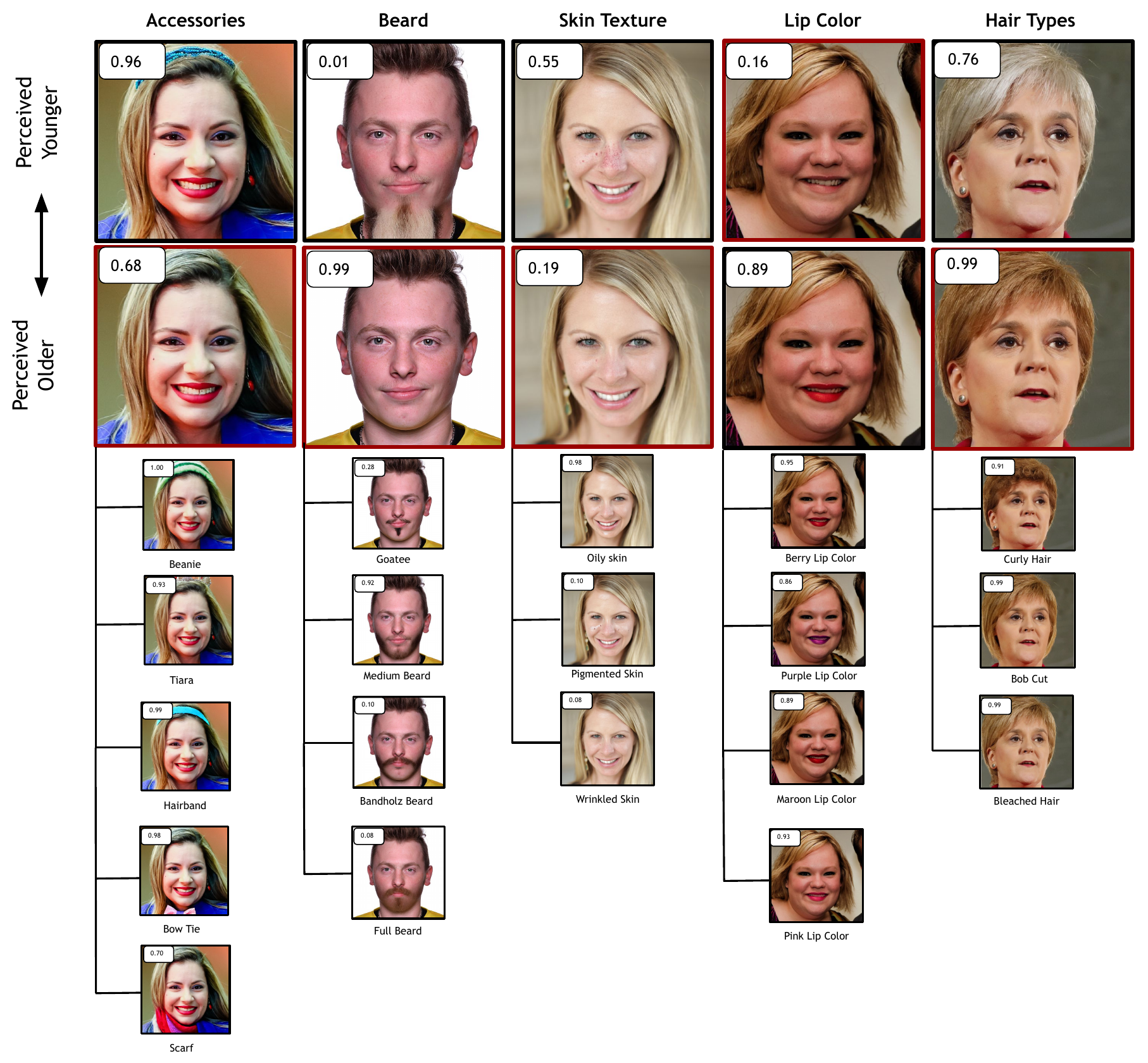}
    \caption{\textbf{Hierarchical Structure of the Top Facial Attributes and their Impact on Age Classifier Scores.} The figure demonstrates how DiffEx organizes fine-grained attribute categories and their influence on classifier decisions. Logit scores in the top-left corners represent the scores for the ``young" label.}
    \label{fig:face_attr}
\end{figure*}

To effectively illustrate the hierarchical structure of the edited features, additional experimental results are presented in Figure \ref{fig:face_attr}, focusing on the facial domain. These results provide a clearer understanding of how specific modifications within different feature categories influence the classifier's output. For instance, as demonstrated in the illustrations, distinct subtypes within a single category, such as various beard styles (e.g., ``stubble," ``goatee," or ``full beard"), exhibit varied impacts on the classifier's score. This highlights the subtle relationship between fine-grained feature variations and their respective contributions to the classification process, showcasing the importance of understanding these hierarchical relationships for improving model interpretability and performance.
\subsection{Experiments with Joint Attributes}
\begin{figure}
    \centering
    \includegraphics[width=1.0\linewidth]{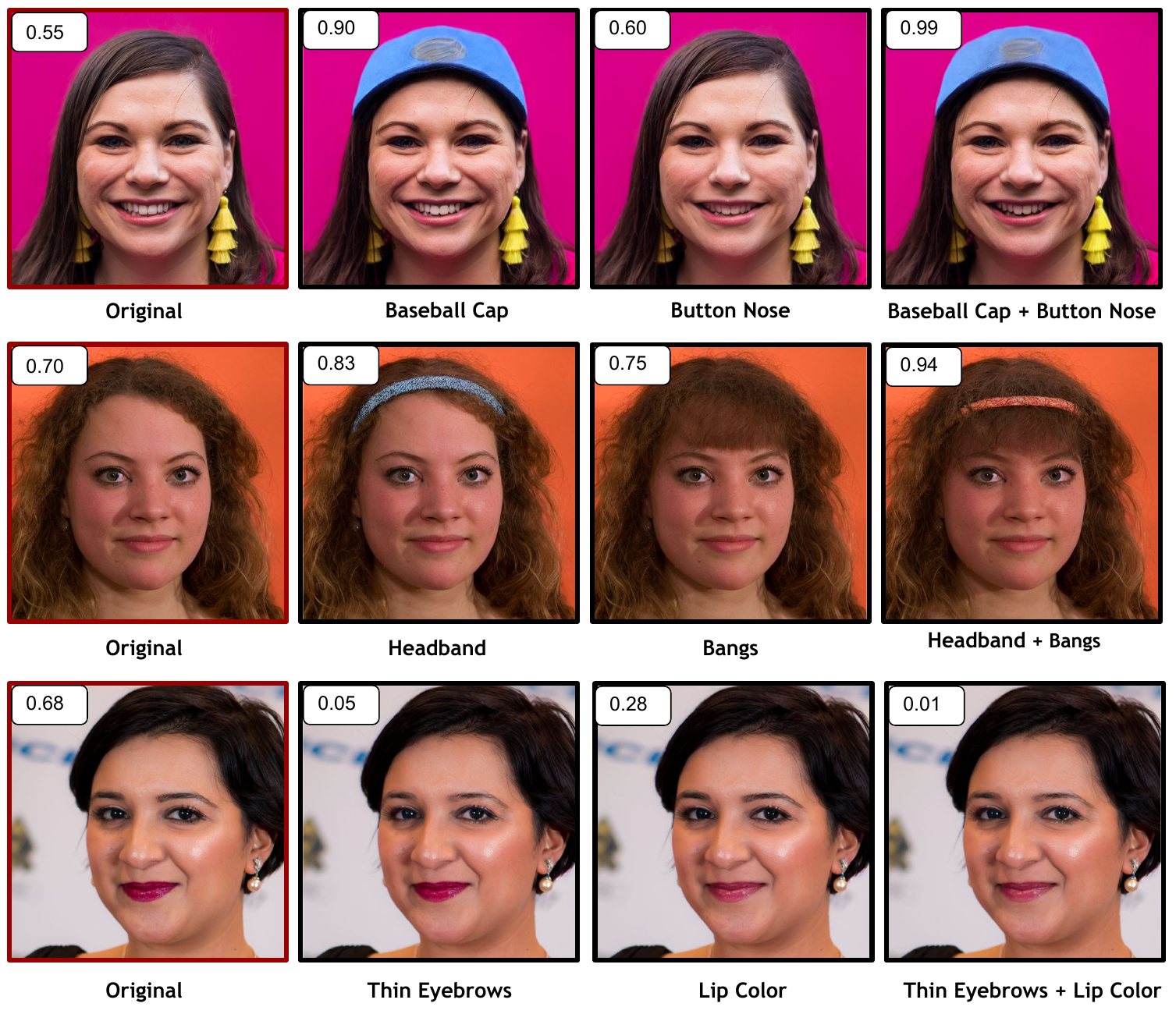}
    \caption{\textbf{Joint Attributes Experiments for the Facial Domain.} This figure showcases some edited facial attributes and their individual and collective effects on the age classifier's decision. The original images, marked with red frames, are compared to their edited counterparts, marked with black frames. The classifier scores displayed in the top-left corner of each image represent how strongly the edited attributes influence the classifier's output. Higher scores indicate a stronger impact of an attribute on a specific domain.}
    \label{fig:joint_face_sup}
\end{figure}

\begin{figure}
    \centering
    \includegraphics[width=1.0\linewidth]{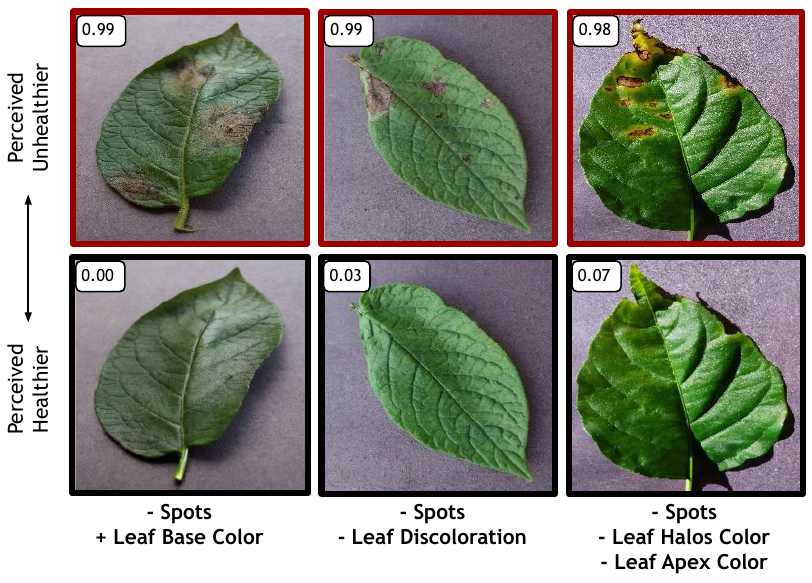}
    \caption{\textbf{Joint Attributes Experiments for the Plant Health Domain.} Here, we present three images edited using joint attributes in the plant health domain. A ``+" sign indicates that an attribute was added to the image, while a ``-" sign signifies that an attribute was removed. Images with red frames represent the original, unedited versions, while those with black frames are the edited versions. The numbers in the corners reflect the classifier's score, indicating the perceived level of the leaf's unhealthiness.}
    \label{fig:joint_leaf}
\end{figure}

\begin{figure}
    \centering
    \includegraphics[width=1.0\linewidth]{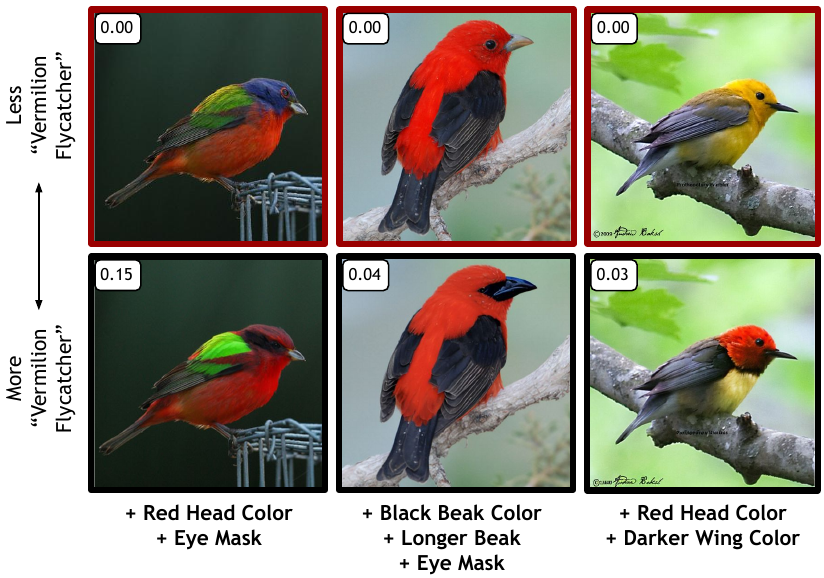}
    \caption{\textbf{Joint Attributes Experiments for the Bird Domain}. Here, we present three images from the bird domain
that were edited to resemble the Vermilion Flycatcher. The focus was on adding attributes to make the birds appear more
similar to the Vermilion Flycatcher. As seen in the joint attributes
figure for the plant health domain, images with red frames represent
the original versions, while those with black frames are the
edited versions. A higher classifier score indicates a greater
resemblance to the Vermilion Flycatcher.}
    \label{fig:joint_bird}
\end{figure}

To examine the impact of combining multiple attributes on classifier scores, we conducted a series of experiments. Specifically, we generated images featuring joint attributes for the face, bird, and plant health classes. The attributes used in these experiments, along with the resulting changes in classifier scores, are presented in Figures \ref{fig:joint_face_sup}, \ref{fig:joint_leaf}, and \ref{fig:joint_bird}.
\subsection{Improving Classifier Accuracy with Counterfactual Images}
After generating counterfactual images for the face domain, we integrated them into the training dataset of image classifiers designed to predict one's gender and age, with the goal of improving their accuracy. The experiments in Table \ref{tab:classifierscores} demonstrate how the classifier's performance changes when 100 counterfactuals containing the ``bangs" and ``makeup" attributes are added to the training data. The original classifiers were convolutional neural networks based on EfficientNet and trained with 1000 images from the FFHQ dataset. Both classifiers achieved an overall accuracy of 95 percent on their test sets. Compared to the other domains, we decided to retrain a classifier with counterfactual images of edited human faces because these images maintain contextually relevant attributes that align with the real-world variations that a classifier will encounter. On the other hand, counterfactuals of edited birds do not reflect realistic bird species (although they can help identify which features of a bird are significant for its overall classification). Thus, these types of edited images introduce features and contexts that are far removed from the target domain, making them unsuitable for training.
\begin{table}[t!]
    \centering
    \renewcommand{\arraystretch}{1.3}
    \begin{tabular}{p{1.5cm} c c c}
        \hline
        \textbf{Attribute} & \textbf{Classifier Type} & \textbf{Original} & \textbf{Updated} \\ \hline
          \textbf{Makeup} & Gender Classifier & 91\% & \textbf{95\%}\\ 
        \textbf{Bangs} & Age Classifier & 68\% & \textbf{96\%}\\ \hline
    \end{tabular}
    \caption{\textbf{Improvement in Classifier Accuracy.} The table illustrates the improvement in the average accuracy of two classifiers in predicting the ages and genders of individuals with makeup and bangs, following the inclusion of counterfactual examples in the face dataset. The ``Original" column presents the average classification scores for individuals with makeup and bangs before the incorporation of counterfactual examples, while the ``Updated" column shows the improved average accuracy scores after adding counterfactual examples to the training data.}
    \label{tab:classifierscores}
\end{table}
\begin{table*}[ht]
\centering
\begin{tabular}{|p{0.2\textwidth}|p{0.3\textwidth}|p{0.45\textwidth}|}
\hline
\textbf{Level 1 Attributes} & \textbf{Level 2 Attributes} & \textbf{Level 3 Attributes} \\ \hline
Face Features & Face Shape, Beard, Mustache & Oval Face, Round Face, Square Face, Heart-Shaped Face, Rectangular Face, Diamond-Shaped Face, Oblong Face, Triangular Face, Long Face, Narrow Face, Wide Face, Broad Face, Full Face, Chunky Face, Wide-Set Face, Expansive Face, Larger Face, Flatter Face, Goatee Beard, Full Beard, Short Beard, Long Beard, Classic Mustache, Handlebar Mustache, Horseshoe Mustache, Pencil Mustache, ... \\ \hline
Hair Features & Hair Color, Hair Texture, Hair Length, Hair Style & Black Hair, Brown Hair, Blonde Hair, Red Hair, Gray Hair, White Hair, Auburn Hair, Straight Hair, Wavy Hair, Curly Hair, Pixie Cut Hair, Bob Cut Hair, Bangs, Permed Hair, Bleached Hair, ...
 \\ \hline
Eyebrow Features & Eyebrow Shape, Eyebrow Density, Eyebrow Style & Arched Eyebrows, Straight Eyebrows, Thick Eyebrows, Thin Eyebrows, Curved Eyebrows, Flat Eyebrows, Angled Eyebrows, Sparse Eyebrows, Dense Eyebrows, Brushed-Up Eyebrows, Plucked Eyebrows, Threaded Eyebrows, ...
 \\ \hline
Mouth Features & Mouth Shape, Lip Volume, Lip Color, Smile Type & Full Lips, Thin Lips, Thick Lips, Wide Mouth, Narrow Mouth, Pouty Lips, Red Lip, Pink Lip, Nude Lip, Coral Lip Color, Berry Lip Color, Brown Lip Color, Purple Lip, Orange Lip, Maroon Lips, ... \\ \hline
Eyelash Features & Eyelash Length, Eyelash Volume, Eyelash Curl & Short Eyelashes, Medium Eyelashes, Long Eyelashes, Sparse Eyelashes, Dense Eyelashes, Straight Eyelashes, Curled Eyelashes, ... \\ \hline
Nose Features & Nose Shape, Nose Tip, Nostril Shape & Straight Nose, Curved Nose, Button Nose, Hooked Nose, Flat Nose, 
Wide Nose, Narrow Nose, Upturned Nose, Long Nose, Broad Nose, Pointed Nose, Roman Nose, Snub Nose, Aquiline Nose, Crooked Nose, Rounded Tip Nose, Pointed Tip Nose, Wide Nostril, Narrow Nostril, Flared Nostril, ...
 \\ \hline
Skin Features & Skin Texture, Skin Color & Smooth Skin, Rough Skin, Oily Skin, Dry Skin, Combination Skin, Sensitive Skin, Acne-Prone Skin, Wrinkled Skin, Freckled Skin, Blemished Skin, Porous Skin, Flaky Skin, Fair Skin, Light Skin, Medium Skin, Dark Skin, Olive Skin, Tan Skin, ...
 \\ \hline
Accessories & Jewelry, Facewear, Headwear & Earrings, Necklace, Bracelet, Ring, Glasses, Sunglasses, Face Mask, Hat, Scarf, Headband, Bow Tie, Hairband, Beanie, Beaded Headband, Tiara, ... \\ \hline
Makeup & Makeup Style, Makeup Type & Natural Makeup, Glam Makeup, Smoky Eye Makeup, Dewy Makeup, Matte Makeup, Bold Lip Makeup, Bridal Makeup, Festive Makeup, Eyeshadow Makeup, Eyeliner Makeup, Blush Makeup, Lipstick Makeup, Highlighter Makeup, Mascara Makeup, ...
 \\ \hline
\end{tabular}
\caption{\textbf{Examples of Attribute Candidates Proposed for the Face Domain.} The table above shows potential level 1, level 2, and level 3 attributes for the face domain. Due to limited space, we include a sample list of level 3 attributes for the first level 2 attribute listed in each row.}
\label{tab:attributes}
\end{table*}
\begin{table*}[ht]
\centering
\begin{tabular}{|p{0.2\textwidth}|p{0.75\textwidth}|}
\hline
\textbf{Level 1 Attributes} & \textbf{Level 2 Attributes}  \\ \hline
Leaf Base Color & Green, Yellow, Light Green, Dark Green, Orange, Red, Brown, Purple, Pink, White, Light Yellow, Dark Red, Burgundy, Copper, Chartreuse, Ivory, Olive, Black, Tan, ... \\ \hline
Leaf Apex Color & Green, Yellow, Red, Purple, Brown, Orange, Pink, White, Light Green, Dark Green, Light Yellow, Dark Red, Rust, Burgundy, Violet, Lime Green, Chartreuse, Copper, Amber, Ivory, ... \\ \hline
Leaf Spots & With Spots, Without Spots \\ \hline
Leaf Disease & Spots, Lesions, Discoloration, Necrosis, Blight, Mold, Mildew, Rust, Canker, Wilting, Decay, Yellowing, Browning, Pustules, Fungal Infection, Bacterial Infection, Viral Infection, Chlorosis, Fungal Growth, Powdery Mildew, Downy Mildew, ... \\ \hline
Leaf Blight Size & Small Blight, Medium Blight, Large Blight, Tiny Blight, Extensive Blight, Minor Blight, Moderate Blight, Severe Blight, Pinpoint Blight, Patchy Blight \\ \hline
Leaf Spots Color & Brown Spots, Yellow Spots, Black Spots, Red Spots, Orange Spots, Green Spots, White Spots, Purple Spots, Light Green Spots, Dark Brown Spots, ... \\ \hline
Leaf Shape & Oblong Shape, Ovate Shape, Lanceolate Shape, Cordate Shape, Elliptical Shape, Linear Shape, Palmate Shape, Pinnate Shape, Lobed Shape, Tamarisk Shape, Sagittate Shape, Triangular Shape, Denticulate Shape, Wedge Shape, Reniform Shape, Setaceous Shape, Circinate Shape, Falcate Shape, Acicular Shape, Subulate Shape, ... \\ \hline
Leaf Symmetry & Bilateral Symmetry, Radial Symmetry, Asymmetrical, Mirror Symmetry, Transverse Symmetry, Rotational Symmetry, ... \\ \hline
\end{tabular}
\caption{\textbf{Examples of Attribute Candidates Proposed for the Plant Health Domain.} The table above lists potential attributes for the plant health domain. However, not all of these attributes are relevant for describing a leaf or would result in effective edits. Therefore, DiffEx filters this list, selecting only the most meaningful and applicable attributes.}
\label{tab:attributes2}
\end{table*}
\begin{table*}[ht]
\centering
\begin{tabular}{|p{0.2\textwidth}|p{0.2\textwidth}|p{0.525\textwidth}|}
\hline
\textbf{Level 1 Attributes} & \textbf{Level 2 Attributes} & \textbf{Level 3 Attributes} \\ \hline
Beak & Beak Color, Beak Shape, Beak Size & Yellow, Orange, Black, Red, Brown, Pink, White, Blue, Green, Grey, Ivory, Cream, Purple, Beige, Tan, Light Pink, Dark Brown, Light Yellow, Dark Green, ... \\ \hline
Wings & Wing Shape, Wing Color, Wing Pattern & Pointed, Rounded, Elongated, Broad, Narrow, Oval, Triangular, Crescent, Oval-Shaped, Square, Short, Long, Fan-Shaped, Forked, Tapered, Slender, Angular, Spade-Shaped, Elliptical, High-Speed, Soaring, High-Aspect Ratio, Cambered, Alula, Swept-Back, V-Shaped, Bent, ...
 \\ \hline
Eye & Eye Shape, Eye Size, Eye Color & Round, Oval, Almond, Circular, Slit, Horizontal, Vertical, Hooded, Wide, Narrow, Protruding, Sunken, Large, Small, Bulging, Beady, Piercing, Squinted, Deep-Set, Prominent, ...
 \\ \hline
Head & Head Color, Crest Presence & Black, White, Yellow, Red, Blue, Brown, Green, Grey, Orange, Pink, Purple, Cream, Beige, Tan, Violet, Charcoal, Silver, Rust, Burgundy, Golden, Copper, ... \\ \hline
Body & Feather Texture, Upperparts Color, Body Size, Belly Color, Tail Length, Leg Color & Soft, Coarse, Smooth, Rough, Fluffy, Silky, Woolly, Feathery, Stiff, Shiny, Matted, Glossy, Velvet, Harsh, Prickly, Fuzzy, Curled, Frizzy, Downy, Crisp, ... 
 \\ \hline
\end{tabular}
\caption{\textbf{Examples of Attribute Candidates Proposed for the Bird Domain.} The table above shows potential level 1, level 2, and level 3 attributes for the bird domain. Due to limited space, we include a sample list of level 3 attributes for the first level 2 attribute listed in each row.}
\label{tab:attributes3}
\end{table*}
\begin{table*}[ht]
\centering
\begin{lstlisting}
[
    {"role": "system",
     "content": 'You are an expert at finding features important for text-based 
     image editing using diffusion models, given a set of images. Upon receiving 
     a set of images, analyze the given inputs and extract important features and 
     keywords that can be used for text-based image editing using diffusion models. 
     Analyze the set of images and identify key features that define or are significant 
     within the specified domain. These features are encoded to guide generative 
     diffusion model for fine-grained image editing of subjects.
     List all different categories related to that specific feature. For example, for DOMAIN_NAME features, it 
     ranges from ATTRIBUTE_1 to ATTRIBUTE_2, ATTRIBUTE_3, ATTRIBUTE_4, etc. 
     Output must be in the format given, a sample output is given below, give the output 
     only without any other descriptive text. Do not restrict your answers to the given 
     sample, come up with all features. I want detailed fine-grained features.
[{
        "ATTRIBUTE_1": {"sub_attribute_1_1" , "sub_attribute_1_2", "sub_attribute_1_3",}
        "ATTRIBUTE_2": {"sub_attribute_2_1", "sub_attribute_2_2", "sub_attribute_2_3"},
        "ATTRIBUTE_3": {"sub_attribute_3_1", "sub_attribute_3_2"},
        "ATTRIBUTE_4": {"sub_attribute_4_1", "sub_attribute_4_2"},
        "ATTRIBUTE_5": {"sub_attribute_5_1", "sub_attribute_5_2", "sub_attribute_5_3"},
}]
\end{lstlisting}
\caption{\textbf{Prompt Template for Keyword-Extraction.} The text above illustrates the standard format used to input text prompts into GPT-4 for extracting potential attributes across different domains. ``DOMAIN\_NAME" refers to a specific domain, such as facial features, bird species, etc. ``ATTRIBUTE\_1, ATTRIBUTE\_2, etc." refer to the Level 1 (broad) categories, while ``sub\_attribute\_1\_1, sub\_attribute\_1\_2, etc." refer to Level 2 (finer-grained) categories.}
\label{tab:gpttemplate}
\end{table*}
\begin{table*}[ht]
\centering
\begin{lstlisting}
[
    {"role": "system",
     "content": 'You are an expert at finding features important for text-based 
     image editing using diffusion models, given a set of images. Upon receiving 
     a set of images, analyze the given inputs and extract important features and 
     keywords that can be used for text-based image editing using diffusion models. 
     Analyze the set of images and identify key features that define or are significant 
     within the specified domain. These features are encoded to guide generative 
     diffusion model for fine-grained image editing of subjects.
     List all different categories related to that specific feature. For example, for human features, it 
     ranges from skin texture to expression, accessories, eyebrow shape, etc. 
     Output must be in the format given, a sample output is given below, give the output 
     only without any other descriptive text. Do not restrict your answers to the given 
     sample, come up with all features. I want detailed fine-grained features.
[{
        "Face": {"oval face" , "rectangular face", "round face",}
        "Skin Texture": {"smooth skin", "freckled skin", "blemish skin", "scar skin"},
        "Skin Color": {"light colored skin", "dark colored skin"},
        "Eyes Shape": {"round eyes", "almond eyes"},
        "Eyes Color": {"blue colored eyes", "green colored eyes", "hazel colored eyes"},
        "Eyebrows": {"thin eyebrows", "bushy eyebrows"},
        "Hair Color": {"dark colored hair", "light colored hair", "blonde hair", 
        "brunette hair",},
        "Hair Texture": {"straight hair", "curly hair", "wavy hair",},
        "Hair Length": {"short hair", "long hair", "medium hair"},
        "Nose Shape": {"button nose", "straight nose", "prominent nose",},
        "Mouth Shape": {"full lip", "thin lip"},
        "Lip Color": {"matte lip", "glossy lip",}
         "earrings", "necklace, glasses, sunglasses",
}]
\end{lstlisting}
\caption{\textbf{Face Domain Keyword-Extraction Prompt Used in GPT-4.} The text above shows the prompt we fed into the VLM in order to find potential attributes in the face domain.}
\label{tab:gptface}
\end{table*}
\clearpage

\end{document}